\documentclass{article}

\PassOptionsToPackage{numbers, compress}{natbib}
\usepackage[preprint]{neurips_2023}

\usepackage[utf8]{inputenc} % allow utf-8 input
\usepackage[T1]{fontenc}    % use 8-bit T1 fonts
\usepackage[hidelinks]{hyperref}       % hyperlinks
\usepackage{url}            % simple URL typesetting
\usepackage{booktabs}       % professional-quality tables
\usepackage{amsfonts}       % blackboard math symbols
\usepackage{amsmath}
\usepackage{nicefrac}       % compact symbols for 1/2, etc.
\usepackage{microtype}      % microtypography
\usepackage[table]{xcolor}  % colors
\usepackage{graphicx}
\usepackage{floatrow}
\usepackage{multirow}       % multirow
\usepackage{subcaption}     % subcaption
\usepackage{makecell}       % for multi-row cells in tables
\newfloatcommand{capbtabbox}{table}[][\FBwidth]
\usepackage{algorithm}
\usepackage{algorithmicx}
\usepackage[noend]{algpseudocode}

\usepackage{xspace}
\newcommand{\algname}[1]{{\sc #1}\xspace}
\usepackage{diagbox}
\usepackage{tabularx}
\usepackage{tikz,tikzscale}  % for tikz figure
\usetikzlibrary{shapes.arrows}

\renewcommand{\paragraph}[1]{ \noindent \textbf{#1}}
\title{\vspace{-4px}SpQR: A Sparse-Quantized Representation for Near-Lossless LLM Weight Compression\vspace{-4px}}

\author{%
  Tim Dettmers\thanks{Equal contribution} \thanks{Corresponding author: \texttt{dettmers@cs.washington.edu}}\\
  University of Washington \\
  \And
  Ruslan Svirschevski\footnotemark[1] \\
  HSE University \& Yandex \\
  \And
  Vage Egiazarian\footnotemark[1] \\
  HSE University \& Yandex \\
  \And
  Denis Kuznedelev\footnotemark[1] \\
  Yandex \& Skoltech \\
  \And
  Elias Frantar \\
  IST Austria \\
  \And
  Saleh Ashkboos \\
  ETH Zurich \\
  \And
  Alexander Borzunov \\
  HSE University \& Yandex \\
  \And
  Torsten Hoefler \\
  ETH Zurich \\
  \And
  Dan Alistarh \\
  IST Austria \& NeuralMagic 
}

\begin{document}

\maketitle

\vspace{-8px}\begin{abstract}

    \vspace{-2px}Recent advances in large language model (LLM) pretraining have led to high-quality LLMs with impressive abilities. 
    By compressing such LLMs via quantization to 3-4 bits per parameter, they can fit into memory-limited devices such as laptops and mobile phones, enabling personalized use. 
    However, quantization down to 3-4 bits per parameter usually leads to moderate-to-high accuracy losses, especially for smaller models in the 1-10B parameter range, which are well-suited for edge deployments. 
    To address this accuracy issue, we introduce the Sparse-Quantized Representation (SpQR), a new compressed format and quantization technique which enables for the first time \emph{near-lossless} compression of LLMs across model scales, while reaching similar compression levels to previous methods. 
    SpQR works by identifying and isolating \emph{outlier weights}, which cause particularly-large quantization errors, and storing them in higher precision, while compressing all other weights to 3-4 bits, and achieves relative accuracy losses of less than $1\%$ in perplexity for highly-accurate LLaMA and Falcon LLMs. This makes it possible to run 33B parameter LLM on a single 24 GB consumer GPU without any performance degradation at 15\% speedup
    %faster inference speed,
    thus making powerful LLMs available to consumer without any downsides. SpQR comes with efficient algorithms for both encoding weights into its format, as well as decoding them efficiently at runtime\footnote{ \href{https://github.com/Vahe1994/SpQR}{\texttt{github.com/Vahe1994/SpQR}}; to be integrated into \href{https://github.com/TimDettmers/bitsandbytes}{\texttt{github.com/TimDettmers/bitsandbytes}}}. Specifically, we provide an efficient GPU inference algorithm for SpQR which yields faster inference than 16-bit baselines at similar accuracy, while enabling memory compression gains of more than 4x.

    % a state-of-the-art quantization technique which identifies and isolates outliers that cause particularly large quantization errors while all non-outlier are quantized to 3-4 bits. With SpQR, we can perform \emph{near-lossless} quantization of LLMs with less than 4.75 bits per parameter such that we maintain a perplexity within 1\% of 16-bit 7-65B parameter LLaMA and Falcon LLMs. When controlling for bits per parameter SpQR significantly improves upon previous methods such as GPTQ. SpQR is based on insights gained from analysing the structure of outlier weights in the quantization process which shows both partially structured outliers (rows, columns, attention heads), as well as unstructured outliers. SpQR makes use of these structures through a 3-4 bit bilevel quantization scheme with a small group sizes of 8-16 weights per group and 16-bit sparse outliers. We develop a fast GPU inference algorithm for SpQR that yields faster inference than 16-bit baselines.

\end{abstract}

%\vspace{-1.5em}
\section{Introduction} 
%\vspace{-0.5em}

Pretrained large language models (LLMs) improved rapidly from task-specific performance \citep{wang2018glue, devlin2018bert, radford2019language}, to performing well on general tasks if prompted with instructions \citep{brown2020language,wei2021finetuned,openai2023gpt}. 
While the improved performance can be attributed to scaling in training data and parameters \citep{kaplan2020scaling,chowdhery2022palm} recent trends focused on smaller models trained on more data, that are easier to use at inference time~\citep{hoffmann2022training, biderman2023pythia, touvron2023llama}. For example, the 7B parameter LLaMA model trained on 1T tokens achieved an average performance only slightly lower than GPT-3 \citep{brown2020language} despite being 25x smaller. Current techniques for LLM compression can shrink these models further by a factor of about 4x, while preserving their performance \citep{dettmers2022llm,xiao2022smoothquant,frantar2022gptq,dettmers2022case}. This yields performance levels comparable to the largest GPT-3 model, with major reductions in terms of memory requirements. With such improvements, well-performing models could be efficiently served on end-user devices, such as laptops.  

The main challenge is to compress models enough to fit into such devices while also preserving generative quality. 
Specifically, studies show that, although accurate, existing techniques for 3 to 4-bit quantization still lead to significant accuracy degradation~\cite{dettmers2022case, frantar2022gptq}. 
Since LLM generation is sequential, depending on previously-generated tokens, small relative errors can accumulate and lead to severely corrupted outputs. 
To ensure reliable quality, it is critical to design low-bitwidth quantization that does not degrade predictive performance compared to the 16-bit model.

In this work, we introduce Sparse-Quantized Representations (SpQR), 
a hybrid sparse-quantized format which can compress accurate pretrained LLMs to 3-4 bits per parameter while staying \emph{near-lossless}: 
specifically, SpQR is the first weight quantization method which is able to reach such compression ratios while 
inducing end-to-end accuracy error as measured in perplexity of less than 1\% relative to the dense baseline. 
SpQR works by combining two innovations. 
First, we isolate \emph{outlier weights}, whose quantization we show to induce disproportionately high errors: these weights are kept in high precision, while the other weights are stored in a much lower, e.g. 3-bit, format.
Second, we implement a variant of grouped quantization with very small group size, e.g. 16 contiguous elements, 
but we show that one can quantize the quantization scales themselves to a 3-bit representation.

To convert a given pretrained LLM into SpQR format, 
we adopt an extended version of the post-training quantization (PTQ) approach recently introduced by GPTQ~\citep{frantar2022gptq}. 
Specifically, the method passes calibration data through the uncompressed model; 
to compress each layer, it applies a layer-wise solver with respect to the L2 error between the outputs of the uncompressed model, and those of the quantized weights. 
% This optimization is done by leveraging Hessian information to influence the rounding decisions, 
% such that the L2 norm of the compression error between the two layer variants is minimized. 
Our approach splits this process into two steps: an ``outlier detection'' step, in which we isolate weights whose direct quantization has outsize impact on layer output behavior, and an actual compression step, in which most ($\geq 99\%$) of weights are compressed to low-bitwidth, the outliers are extracted, and the whole representation is rendered more efficient by further compressing the quantization metadata. 

% starts by observing that the GPTQ approach can be re-interpreted as providing an ordering of the weights with respect to their induced quantization error. 
% Then, we observe that these errors tend to be \emph{well-structured}, in the sense that certain contiguous groups of weights have an outsize contribution towards the total L2 error. 

Our method is motivated by a new analysis showing that LLM weight quantization errors exhibit both vertical and horizontal group correlations, corresponding to systematic large errors corresponding to input feature dimensions and output hidden dimensions. While outlier input features have been observed before~\citep{dettmers2022llm,xiao2022smoothquant}, our work is the first to demonstrate that similar outliers occur \emph{in the weights, for particular output hidden dimensions}. Unlike input feature outliers, the output hidden dimension outliers occur only in small segments for a particular output hidden dimension. 

Our quantization algorithm isolates such outliers and efficiently encodes a given model in SpQR format. 
To exploit the resulting structure, we develop a specialized sparse-matrix multiplication algorithm based on the compressed sparse row (CSR) format. To use SpQR for token-by-token generation, we combine this sparse algorithm together with a dense-quantized matrix multiplication for 3-4 bit weights. With this, SpQR reduces the memory footprint of LLMs by a factor of about 3.4x or more without degradation in accuracy, measured as language modeling loss or perplexity, while also being 20-30\% faster for LLM generation compared to 16-bit inference. 
% Thus, SpQR paves the way for \emph{highly-accurate, efficient compressed execution of LLMs} on memory-constrained devices.

\begin{figure}[t!]
    \vspace{-0.5em}
    \centering
    \begin{subfigure}{0.47\linewidth}
        \includegraphics[width=\linewidth]{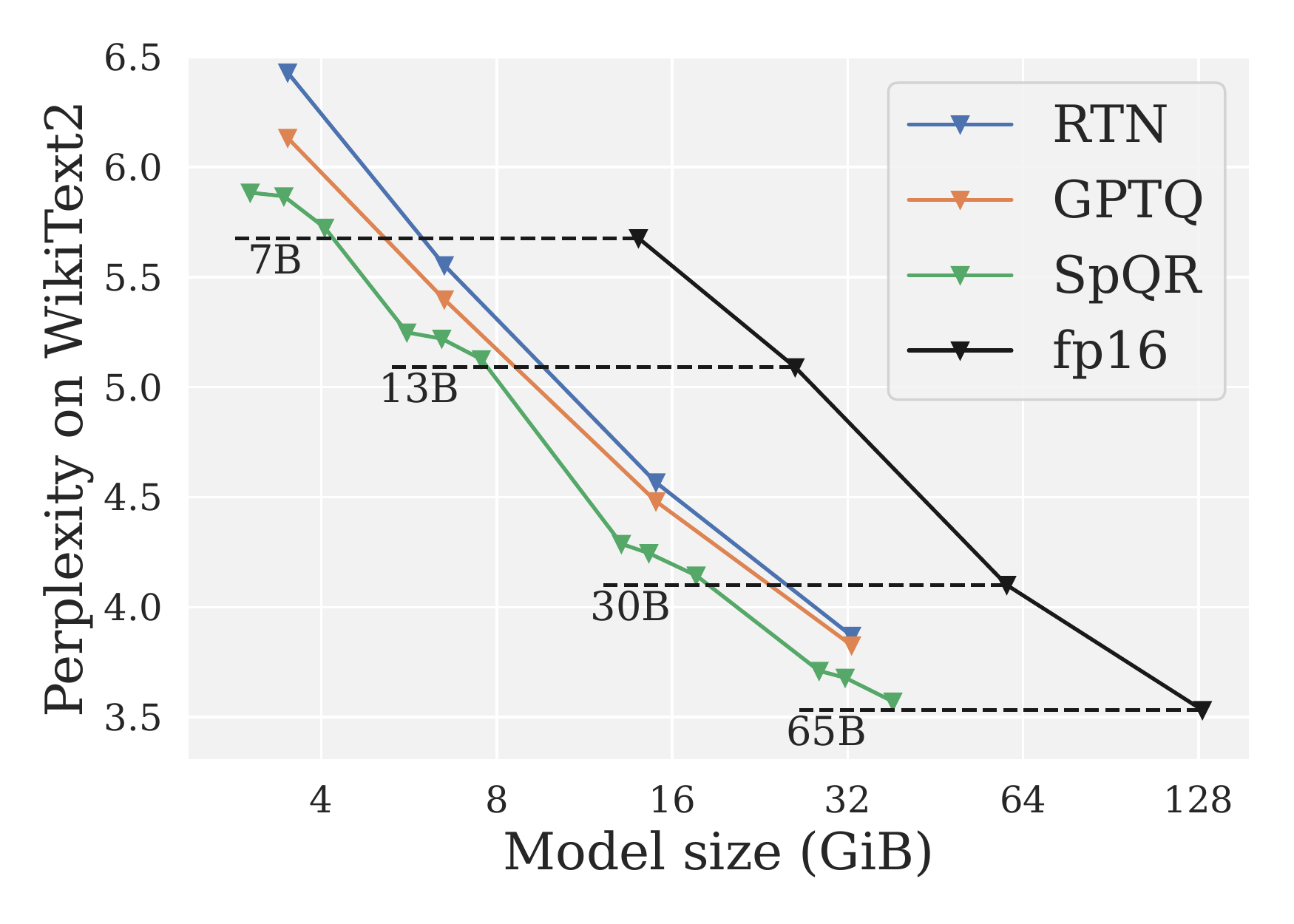}
    \end{subfigure}
       \begin{subfigure}{0.47\linewidth}
        \includegraphics[width=\linewidth]{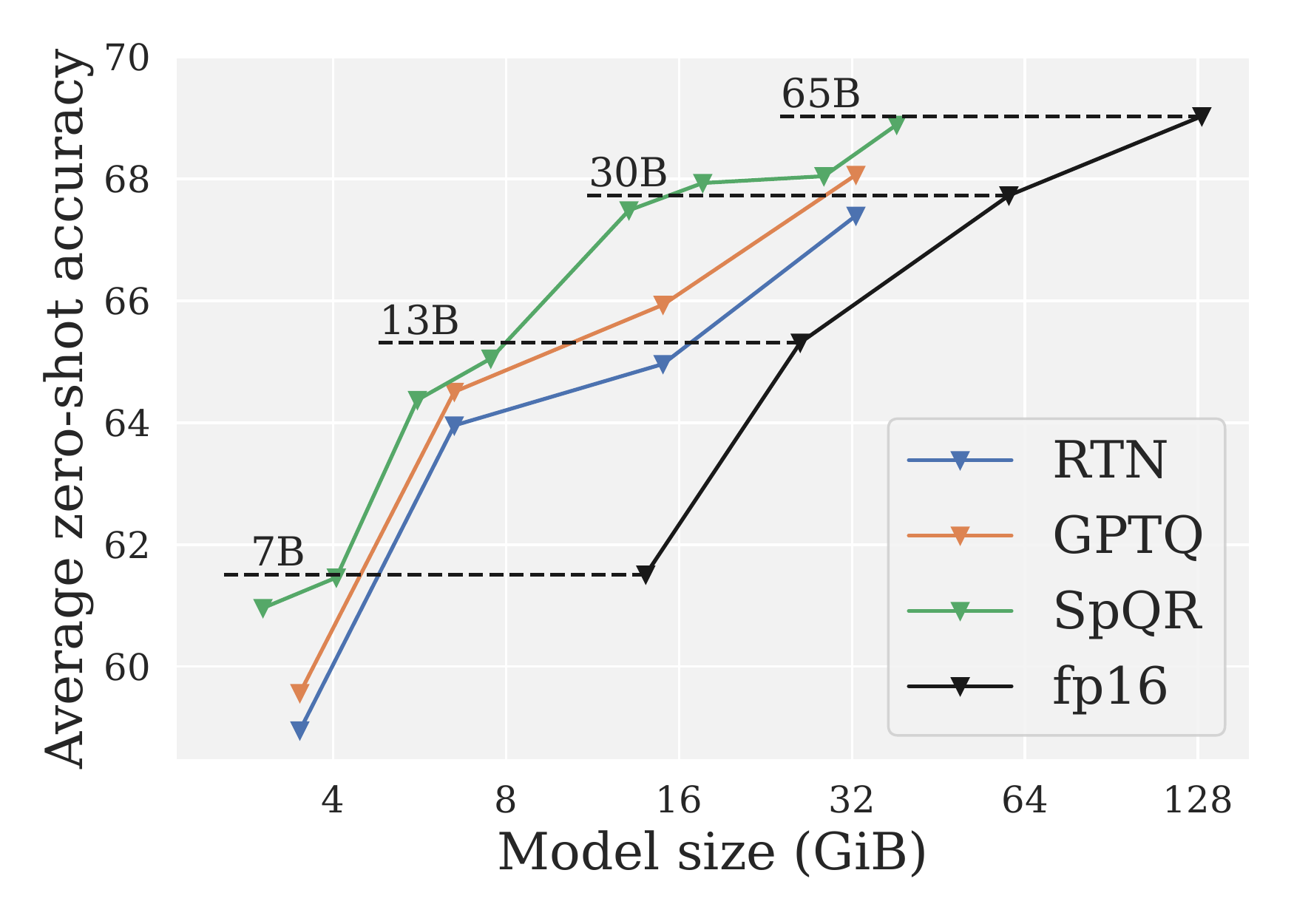}
    \end{subfigure}
    \vspace{-8px}
    \caption{
       Compressed LLM performance for LLaMA models. % Denis, shall i keep this sentence or do you have other plans?
       (\textbf{left}) LM loss on WikiText2 vs model size.
       (\textbf{right}) Average performance on zero-shot tasks vs model size. 
    }
    \label{fig:quantization_method_comparison}
\end{figure}

% \section{Introduction}\label{sect:intro}

% \begin{itemize}
%     \item Introduction about promise and challenges of LLMs.
%     \item Focus on generative mode, motivating weight size as a challenge to edge deployment. 
%     \item Overview of existing techniques and remaining challenges.
%     \item Overview of contribution. 
% \end{itemize}

% \section{Background}\label{sect:background}

% \vspace{-0.7em}
\section{Related Work}
% \vspace{-0.5em}

We focus our discussion on related \emph{post-training quantization (PTQ) methods}~\citep{nagel2020up}, referring the reader to the recent survey of Gholami et al.~\citep{gholami2021survey} 
for full background on quantization. 
PTQ methods are a popular approach for \emph{one-shot compression} of models with various sizes, based on a limited amount of calibration data, 
using accurate solvers, usually focused on layer- or group-wise compression sub-problems.
Most PTQ methods, such as AdaRound~\citep{nagel2020up}, BitSplit~\citep{wang2020towards}, AdaQuant~\citep{hubara2021accurate}, BRECQ~\citep{li2021brecq}, or OBQ~\citep{frantar2022obc}  were designed for vision models or small-scale language models, with less than 100M parameters. 
% In this line of work, proposed a customized solver inspired by the Optimal Brain Surgeon approach~\cite{lecun1990optimal, hassibi1993optimal} for minimizing the squared error of the layer-wise quantization problems. 
All these recent approaches tend to use accurate solvers, which would not scale to GPT-scale models in terms of computational or memory cost, as they are 10-1000x larger in size.   

Recently, there has been significant interest in obtaining accurate post-training methods that scale to such massive models. 
Due to computational constraints, early work such as ZeroQuant~\citep{yao2022zeroquant}, LLM.int8()~\citep{dettmers2022llm}, and nuQmm~\citep{park2022nuqmm} 
used direct rounding of weights to the nearest quantization level, while customizing the quantization granularity (i.e., group size) to trade off space for increased accuracy. LLM.int8()~\citep{dettmers2022llm} suggested isolating ``outlier features'' which would be quantized separately to higher bit-width. 
These approaches are able to induce relatively low quantization error, e.g. 5.5\% relative LM Loss increase for LLaMA-7B at 4-bit weight quantization, provided that the quantization granularity is low enough. 
GPTQ~\cite{frantar2022gptq} proposed a higher-accuracy approach (e.g., 4\% LM Loss increase in the above setting), which works via an approximate large-scale solver for the problem of minimizing the layer-wise squared error.  
% GPTQ is able to reach fairly accurate 3-bit quantization for low-enough group size. 

Dettmers et al.~\cite{dettmers2022case} provided an in-depth overview of the accuracy-compression trade-offs underlying these methods, 
establishing that 4-bit quantization is an optimal point for round-to-nearest-based methods,  
whereas higher compression can be achieved via data-aware methods such as GPTQ. 
SparseGPT~\cite{frantar2023massive} presented an approach to jointly sparsify LLM weights to medium sparsities, together with quantization of the remaining weights to a fixed given bit-width. 
One common drawback of existing methods is that the accuracy loss relative to the original model is still significant (see Table~\ref{tab:quantization_method_comparison_LLaMA}). 
This is especially relevant to relatively small but easily deployable models, e.g. in the 7-13B parameter range, where existing methods show drastic accuracy drops. 
We investigate this question here, and provide a new compression format which can lead to near-lossless 3-4 bits compression in this regime. 

%The main drawback is that to deploy common high-quality models in the 7-13B parameter range unto memory limited devices a 3-4 bit quantization is needed that incurs a 6-11\% drop in PPL for LLaMA family models. Such degradation in performance can lead to a severe and unpredictable loss in generation quality. As such, it is critial to develop compression schemes that are near-lossless for 3-4 bit quantizations.

A related question is that of performing both activation and weight quantization. 
There is early work~\cite{dettmers2022llm, xiao2022smoothquant, yao2022zeroquant}, 
showing that both activations and weights could be quantized to 8-bits with relatively low accuracy impact. 
These complementary investigations yield interesting insights into the causes of compression error in the case of LLMs. 
Specifically, \cite{dettmers2022llm, xiao2022smoothquant} observe the presence of ``outlier features'' with significantly higher values in the input/output of large LLMs, which induce higher quantization error, and propose different mitigation strategies. 
%We analyze this phenomenon from the point of view of weight quantization, and isolate the presence of ``outlier weight blocks'' from the point of view of the quantization error. 
%Our analysis shows that outlier weight blocks are correlated but do not precisely correspond to outlier features, which motivates the introduction of our fine-grained hybrid compressed weight format. 

We analyze this phenomenon from the point of view of weight quantization. In particular, we investigate the outlier structure, beyond input feature outliers in the weight matrix. While we find that input feature outliers of the current layer are correlated to hidden unit outliers weight in the previous layer there is not a strict correspondence. Such partially-structured outlier patterns necessitate a fine-grained hybrid compression format that goes beyond algorithms that exploit the column structure of outlier features found in previous work.

Hybrid sparse-quantized formats have been investigated generally for deep networks. 
Some efficient CPU inference engines~\cite{deepsparse, gorbachev2019openvino} support a different block sparse-and-quantized format, 
in which each block of $4$ consecutive weights is either completely sparse or quantized to 8-bit format, 
whereas GPUs support a similar compound format in which every group of 4 weights contains 2 zero weights, 
and the non-zero weights could be quantized. 
The FBGEMM package~\cite{fbgemm} proposed a format in which certain ``outlier'' weights are quantized separately, 
to reduce their impact on normalization. 
However, in this format, ``outlier'' weights are still quantized to exactly the same bit-width (8-bit) as regular weights; 
moreover, no procedure is given for converting a model to this format post-training. 
By contrast, 
1) we provide an efficient and accurate post-training compression algorithm which identifies outliers as weights inducing high output error, 
2) we propose a format compressing outliers to a higher bit-width relative to regular weights, and 
3) our format stores outliers in blocks, allowing for efficient implementation of GPU kernels, which we provide as well. 

\vspace{-0.7em}
\section{Quantization sensitivity of LLM weights}
\vspace{-0.5em}
\subsection{Parameter sensitivity under quantization}\label{sect:method_understand}
\vspace{-0.5em}
% Not all parameters of a large language models are equally important. If we can find and isolate the most important weights in higher precision, this could lead a quantization with less degradation. In this section, we take a closer look at LLM weight matrices and determine which weights are more sensitive to compression. In particular, we first describe the intuition behind what a sensitive weight is, then we describe how we find sensitive weights with the GPTQ algorithm, and finally we present an overview of patterns of particularly sensitive weights (outlier weights) that we found by using GPTQ algorithm on LLaMA model weights. These results severe as a basis for motivating our main contribution, the SpQR representation. In our analysis section, we provide additional detail of the outlier weight patterns we found.

% Analyzing weight sensitivity can be deceptive: even if a given parameter has large quantization error, modern quantization algorithms can nullify this error by adjusting other parameters~\cite{gptq,deepquant,obc}. Ideally, we should \textit{seek parameters that, when quantized, introduce large irreducible error to model behavior}.

% NOTE: suggested rewrite below, maybe it's a bit too long; I also tried to make it sound as novel as possible

Not all parameters in a neural network are equally important. 
Intuitively, a weight could be seen as sensitive to quantization if its rounding error is large, i.e. it is not close to a quantization point, and/or the inputs it is usually multiplied with a large, amplifying even a small rounding error. These simple notions of sensitivity however disregard the fact that LLMs operate on very large vectors with significant correlations: a weight $w_a$ may have a large rounding error while being strongly correlated to another weight $w_b$, meaning that the error of rounding up $w_a$ can be well compensated by rounding down $w_b$. This idea is exploited by modern quantization algorithms \cite{frantar2022gptq, yao2022zeroquant} and can lead to major improvements over vanilla rounding, especially a low bitwidths. Properly capturing this aspect of sensitivity requires a more robust definition.

For computational tractability, we assess sensitivity on a per-layer level using a small set of \emph{calibration inputs} $X$, collected by running them through the model up to the particular layer. We define the sensitivity $s_{ij}$ of some weight $w_{ij}$ in the layer's weight matrix $W$ as the minimum squared difference between the original predictions on $X$ and those of any weight matrix $W'$ where this weight is quantized, i.e. $w'_{ij} = \text{quant}(w_{ij})$:
\begin{equation}
    s_{ij} = \text{min}_{W'} \, ||WX - W'X||_2^2 \quad \text{s.t.} \quad w'_{ij} = \text{quant}(w_{ij})
\end{equation}
Crucially, all weights of $W'$ except for $w'_{ij}$ may take on arbitrary, not necessarily quantized, values in order to compensate for the quantization error incurred by rounding $w_{ij}$, thus capturing the correlation aspect discussed above. Further, as we allow continuous values, this problem admits a closed-form solution. This can be determined by following the generalized Optimal Brain Surgeon framework \cite{frantar2022obc}, where $(XX^\top)^{-1}$ is the inverse Hessian matrix corresponding to the optimization problem:
\begin{equation}\label{eq:error}
    s_{ij} = \frac{(w_{ij} - \text{quant}(w_{ij}))^2}{2(XX^\top)^{-1}}.
\end{equation}
This saliency measure can be approximated efficiently by quantization solvers, such as GPTQ~\cite{frantar2022gptq}. In more detail, GPTQ quantizes weight matrices column-by-column while in each step adjusting the not-yet-quantized part to compensate for the quantization error in a similar sense as defined above. Consequentially, instead of statically deciding all sensitivities in advance, they can be computed dynamically as the algorithm processes each column, by using the inverse of the Hessian subselection corresponding to all not yet quantized weights. This matrix is already efficiently computed by GPTQ and thus does not impose any additional overheads. The main advantage of this approach is that $s_{ij}$ is always determined based on the most current value of $w_{ij}$ and thus accounts for adjustments due to previously quantized weights as well.

\subsection{Exploring parameter sensitivity}\label{sect:method_llama65}

\begin{figure}[h!]
    \centering
    \includegraphics[height=150px]{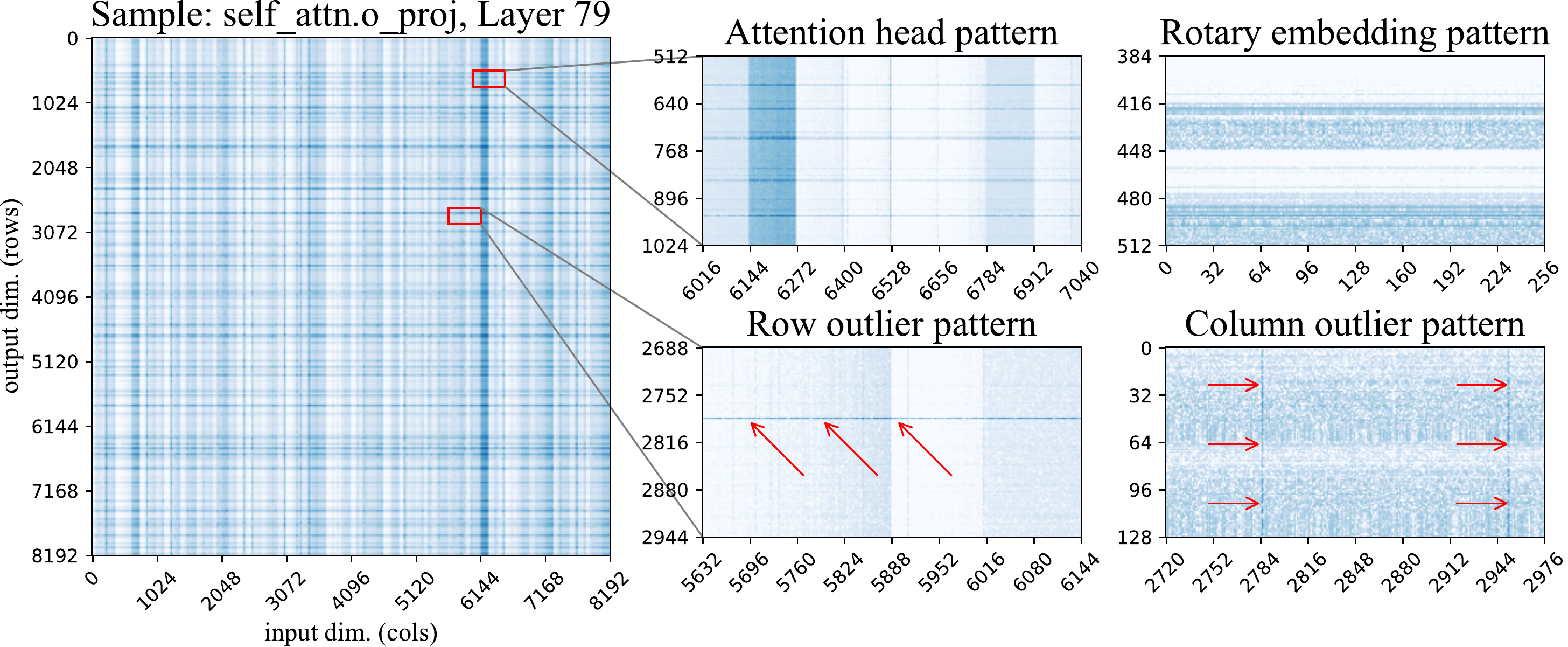}
    \caption{Weight log-sensitivities from the last attention layer of LLaMA-65B. Dark-blue shades indicate higher sensitivity. The image on the left is a high-level view, resized to 1:32 scale with max-pooling. The two images in the middle are zoomed in from the main figure. The two images on the right are taken from other weight matrices.}
    \label{fig:patterns}
\end{figure}

Before we define out main method, SpQR, we provide a motivating analysis of parameter sensitivity which uncovers that the location of sensitive weights in the weight matrix are not random but have particular structures. To highlight these structural elements during the quantization process, we calculate the the per-weight sensitivities and visualize them for the popular and highly-accurate LLaMA-65B model~\cite{touvron2023llama}. 
As the quantization method, we use GPTQ quantization to 3-bit, without weight grouping, following~\citep{frantar2022gptq}. We use C4~\cite{C4} as the calibration dataset, and we estimate the error on 128 sequences of 2048 tokens each. 
Figure~\ref{fig:patterns} depicts the output projection of the last self-attention layer of LLaMA-65B. 

% (In the HuggingFace implementation, this corresponds to \texttt{model.layers.79.self\_attn.o\_proj}.)
Using the sensitivity analysis, we observe several patterns in the weight matrix, often in a single row or column. 
Since the large weight matrices in LLaMA-65B have too many rows/columuns to be respresentable in a compact image (default: 8k $\times$ 32k pixels) we perform max pooling to visualize the matrices, that is we take the maximum sensitivity in each square of $32\times32$ rows and columns. This max pooling only affects the leftmost image.
Using this visualization, we observe that the quantization error patterns vary both by layer type, for example attention vs multilayer perceptron (MLP), and layer depth. In particular, we find that more sensitive outliers are present for deeper layers.  (Please see Appendix \ref{app:extra_analysis} for additional results.) 
We now proceed to categorize outlier structures, taking this  attention weight matrix as an exemplar. We make the following observations:

\begin{itemize}
    \item \textbf{Row outliers} are shown in Figure~\ref{fig:patterns} bottom-center as regions of high sensitivity within one output unit. Some of these patterns span the entire row, while others are partial. In attention layers, some of the partial row outliers correspond to some subset of attention heads. \textbf{Column outliers} appear in Figure~\ref{fig:patterns}, bottom-right, showing high sensitivity in select input dimensions (columns) across all rows. The latter are correlated to the ``outlier feature'' phenomenon reported in Dettmers et al.~\cite{dettmers2022llm}.
    
    \item \textbf{Sensitive attention heads.} (Figure~\ref{fig:patterns}, top-center) -- regular stripes of width 128 highlight all weights corresponding to one attention head.  This could be related to some attention heads having more important functions~\cite{voita-etal-2019-analyzing,vig2019multiscale,olsson2022context}. 
The corresponding ``stripes'' are horizontal for attention Q \& K projections, vertical in output projection, and absent from value projections and any MLP weights. Of note, there is significant variation in individual weight sensitivity even within the sensitive heads.
    \item  
\textbf{The Rotary embedding pattern}, a repeating vertical pattern of sensitivity with a period of 64 units. We attribute this to the use of rotary embeddings~\cite{su2021roformer}: each attention head (dim = 128) is split into two halves: the first 64 are ``rotated'' with cosine, and the other 64 use sine. Both sine and cosine rotation use the same set of frequencies. Typically, the weights that correspond to low-frequency sines and cosines are more sensitive than their high-frequency counterparts, as shown in Figure~\ref{fig:patterns} (top-right). As expected, this pattern is absent from any layer not using rotary embeddings.

    \item 
\textbf{Unstructured outliers.} Besides the above, each layer has a number of individual sensitivity weights that do not fit into any of the above patterns.
These unstructured outliers occur more frequently for columns with largest input index (i.e. on the right side of the images). This effect is difficult to see on a heatmap, so we provide additional figures and statistical tests in Appendix \ref{app:extra_analysis}.
We believe is probably an artefact of the GPTQ algorithm, which compresses one by one, using yet-uncompressed weights to compensate the error. 
Thus, the rightmost batch of weights accumulates the most error.

\end{itemize}

Next, we will leverage these findings to propose a compressed representation which can support all these different outlier types.

% \vspace{-0.7em}
\section{SpQR: A Sensitivity-aware compressed representation}\label{sect:method_compress}
% \vspace{-0.5em}

\subsection{Overview}

Existing LLM quantization algorithms treat low- and high-sensitivity weights equally; however, our above discussion suggests that this may lead to sub-optimal quantization. 
Ideally, we would want the representation to assign more of its ``size budget'' to sensitive weights. However, these weights are scattered in the weight matrix as either individual weights or small groups, for example, partial rows or attention head. To capture this structure, we are introducing two changes to the quantization procedure: one for capturing small sensitive groups, and another for capturing individual outliers.

\textbf{Capturing small groups of weights with bilevel quantization.}
% outliers affect quantization due to reducing the utilization of quantization bins after normalization into the maximum range of the quantization range. With a small group size, we can limit this effect to a small group of weights rather than all weights in the weight tensor. The smaller the group, the more we isolate outliers into particular groups. However, this increases the memory required to store gropps. As such we need develop bilevel quantziation 
In the previous section, we observed several cases where weights behave similarly in small consecutive groups, with abrupt changes between groups, for example for some attention head and partial row outliers (see Figure~\ref{fig:outliers_fig} left, bottom-center). When applying a standard approach, there will be many cases where these weights will be grouped together, sharing the same quantization statistics. To reduce the number of such cases, we use groupwise quantization with extremely small groups, typically of $\beta_1{=}8-32$ weights.
That is, for every $\beta_1$ consecutive weights, there is a separate quantization scale and zero-point. 
This choice runs contrary to current intuition: for instance, the recent work of Yao et al.~\cite{yao2023comprehensive}
explicitly recommends against small groups, arguing that the overhead for storing quantization statistics would outweigh the precision advantages.

To circumvent this issue, we quantize the groupwise statistics themselves using the same quantization algorithm as for weights --- asymmetric (min-max) quantization. Because of how min-max quantization works, the range of quantized values will fit to the groups with largest (or smallest) quantization scale, quantizing them perfectly.
In other words, we group groupwise statistics from $\beta_2=16$ consecutive values and quantize them together in the same number of bits, such that groups with atypical quantization parameters end up using more of the ``quantization budget''. Finally, both first and second-level quantization is directly within the quantization process, allowing the algorithm to compensate the second-level quantization error where possible.

\textbf{High-sensitivity outliers.} Our analysis showed the existence of cases where a small percentage of sensitive weights come in small groups (in the self-attention) or individual ``outliers'' (in the MLP). 
In some cases, 1\% of the weights account for over 75\% of the total quantization error. 
Since these weights appear to lead to high, irreducible error, we choose to keep these outliers in high precision (16-bit).
As these outliers are often unstructured, we encode them individually in a row-wise arrangement similar to a compressed-sparse-row (CSR) representation~\cite{hoefler2021sparsity}. This can encode both individual outliers and small structures that do not fit into the above definition of groups.

The procedure for detecting the outliers is described in detail in 
Alg.~\ref{alg:main}. If follows a rough two-step procedure: (1) find and isolate outliers as 16-bit weights, (2) quantize the non-outlier ``base'' weights into 3-4 bit and transfer the remaining quantization into the the 16-bit outliers weights.
% At a high , this estimates the reduction in the layer's error caused by exempting the given weight from quantization. 
% Please see Alg.~\ref{alg:main} (right) for the analytic formulas and their implementation. 
For the outlier isolation step, the algorithm implements a filtering technique based on the sensitivity criterion in Eq.~\eqref{eq:error},  
which is used to isolate and separate outliers from base weights. 
Globally, for each matrix, the algorithm aims to pick a sensitivity threshold $\tau$ to obtain the desired number of outliers across the whole model, usually around $1\%$ of weights.
Specifically, a particular weight is considered an outlier if keeping the weight in 16-bit reduces the error in Eq.~\eqref{eq:error} by at least $\tau$. 

Following this first outlier detection step, we quantize the base weights ignoring all outliers that occur in the same quantization group. As such, the quantization statistics (e.g. scales) are computed by excluding outliers. This results in significant improvements in terms of error, since e.g. the min-max scales will be significantly reduced.  
The algorithm then proceeds to apply GPTQ to quantize the remaining weights.  
% (We show in Section~\ref{sect:experiments} that the algorithm is robust to the choice of $\tau$.) 
Interestingly, unlike~\cite{dettmers2022llm}, a weight can be chosen to be an outlier not only if it causes error by itself, but also if the GPTQ algorithm can employ this weight to compensate errors from many other weights. Thus, the resulting 16-bit value will contain not the original weight, but a weight that was adjusted to minimize the output error. As such, SpQR goes beyond mere detection of outliers towards the more general notion of isolating and treating outliers that occur {\it during} the quantization process.
Finally, the algorithm gathers and compresses sparse outlier matrix as well as the final quantization statistics with bilevel quantization and returns the compressed weights and their metadata.

\begin{algorithm}
\caption{SpQR quantization algorithm: the left snippet describes the full procedure, the right side contains subroutines for bilevel quantization and finding outliers.}
\label{alg:main}
\small
\begin{minipage}{0.6\textwidth}
    \texttt{func} \algname{SpQRQuantize($W, X, b, \beta_1, \beta_2, \tau, \lambda$)}
    \begin{algorithmic}[1]
      \Require $W\in \mathcal{R}^{m\times n}$ --- weight matrix,
      \Statex \quad\space\space $X\in \mathcal{R}^{n\times d}$ --- calibration data,
      \Statex \quad\space\space $b$ --- the base number of quantization bits,
      \Statex \quad\space\space $\beta_1, \beta_2$ --- quantization group sizes,
      \Statex \quad\space\space $\tau$ --- sensitivity outlier threshold
      \Statex \quad\space\space $\lambda$ --- hessian regularizer,
      \Statex
      \State $E := \text{float\_matrix}(m, n)$ \quad // L2 error
      \State $H := 2 X X^T$  \quad // L2 error hessian, $\mathcal{R}^{n \times n}$
      \State $H^{\text{ic}} := \text{Cholesky}((H + \lambda \mathbf{I})^{-1})$
      \State $Q := \text{int\_matrix}(m, n)$ \quad  // quantized weight
      \State $\mathcal{O} := \emptyset$ \quad  // a set of all outliers
      \State $\mathcal{S} := \emptyset$ \quad  // a set of quantization statistics
      \For{$i = 1, \beta_1, 2 \beta_1, \dots n$}
        \State $W_{:, i: i + \beta_1}, \mathcal{O} := \text{outliers}(W_{:, i: i + \beta_1}, H^{\text{ic}}_{i: (i + \beta_1), i: (i + \beta_1)} \mathcal{O})$
        \State $\hat s, \hat z, \mathcal{S} := \text{fit\_statistics}(W_{:, i: i + \beta_1}, \mathcal{S}, \mathcal{O})$
        \For{$j = i, \dots, i + \beta_1$}
          \State $Q_{:, j} := \text{quantize}(W_{:, j}, \hat s, \hat z)$
          \State $\vec w_q := \text{dequantize}(Q_{:, j}, \hat s, \hat z)$
          \State $E_{:, j} := (W_{:, j} - \vec w_q) / H^{\text{ic}}_{j, j} \cdot (1 - \text{is\_outlier}(W_{:, j}, \mathcal{O}))$
          \State $W_{:, j:(i + \beta_1)} := W_{:, j:(i + \beta_1)} - E \cdot H^{\text{ic}}_{j, j:(i + \beta_1)}$
        \EndFor
        \State $W_{:, (i + \beta_1) : n} := W_{:, (i + \beta_1):n} - E \cdot H^{\text{ic}}_{i: (i + \beta_1), i:(i + \beta_1)}$
      \EndFor
      \State $S_q, Z_q, S_s, Z_s, S_z, Z_z := \text{gather\_statistics}(\mathcal{S})$
      \State $W_{sparse} = \text{gather\_outlier\_matrix}(W, \mathcal{O})$
      \State\Return $Q, S_q, Z_q, S_s, Z_s, S_z, Z_z, W_{sparse}$
    \end{algorithmic}
    \vspace{8px}
    \texttt{func} $\textbf{quantize}(M, \vec s, \vec z)$
    \begin{algorithmic}[1]
    \State\Return $\lfloor M / \vec s + \vec z + 0.5 \rfloor$
    \end{algorithmic}
    \vspace{8px}
    \texttt{func} $\textbf{dequantize}(Q, \vec s, \vec z)$
    \begin{algorithmic}[1]
    \State\Return $\vec s \cdot (Q - \vec z)$
    \end{algorithmic}
\end{minipage}
\hspace{10px}\begin{minipage}{0.35\textwidth}
    \texttt{func} $\textbf{fit\_quantizer}(M, \beta)$
    \begin{algorithmic}[1]
    \State $\vec m := \text{flatten}(M)$
    \State $\vec s, \vec z := \text{vectors()}$
    \For{$i = 1, \beta_1, 2 \beta_1, \dots \text{dim(m)}$}
      \State $s_i := \frac{\text{max}({\vec m}_{i: i + \beta}) - \text{min}({\vec m}_{i: i + \beta})}{2^b - 1}$
      \State $z_i := - \text{min}({\vec m}_{i: i + \beta}) / s_i$
    \EndFor
    \State\Return $\vec s, \vec z$
    \end{algorithmic}

    \texttt{func} $\textbf{error}(W, H^{\text{ic}})$
    \begin{algorithmic}[1]
    \State $\vec s, \vec z := \text{fit\_quantizer}(W, \beta_1)$
    \State $W_q := \text{quantize}(W, \vec s, \vec z)$
    \State $E := (W - W_q) / H^{\text{ic}}$
    \State\Return $E ^ 2$
    \end{algorithmic}
    
    \texttt{func} $\textbf{outliers}(W, H^{\text{ic}}, \mathcal{O})$
    \begin{algorithmic}[1]
    \State $E_{\text{base}} = \text{error}(W, H^{\text{ic}})$
    \For{$i = 1, \dots, \beta_1$}
      \State $loo := \{1, 2, ..., \beta_1\} / \{i\}$
      \State $E_{\text{ol}} = \text{error}(W_{:, \text{loo}}, H^{\text{ic}}_{\text{loo}, \text{loo}})$
      \State $I_o = \text{select}(E_{\text{base}} - E_{\text{ol}} > \tau)$
      \State $\mathcal{O} := \mathcal{O} \cup I_o$    
    \EndFor
    \State\Return $W, \mathcal{O}$
        %     \State $\vec s, \vec z := \text{fit\_quantizer}(W_{:, i: i + \beta_1}, \beta_1)$
        % \State // $\vec s$ for scales, $\vec z$ for zero points
        % \State $\vec s_s, \vec z_s := \text{fit\_quantizer}(\vec s, \beta_2)$
        % \State $\vec s_z, \vec z_z := \text{fit\_quantizer}(\vec z, \beta_2)$
        % \State $\vec s_q := \text{quantize}(\vec{s}, \vec s_s, \vec z_s)$
        % \State $\vec z_q := \text{quantize}(\vec{z}, \vec s_z, \vec z_z)$
    \end{algorithmic}
    
    \texttt{func} $\textbf{fit\_statistics}(W, \mathcal{S}, \mathcal{O})$
    \begin{algorithmic}[1]
        \State $W := W \cdot (1 - \text{is\_outlier}(W, O))$
        \State $\vec s, \vec z := \text{fit\_quantizer}(W, \beta_1)$
        \State // $\vec s$ for scales, $\vec z$ for zero points
        \State $\vec s_s, \vec z_s := \text{fit\_quantizer}(\vec s, \beta_2)$
        \State $\vec s_z, \vec z_z := \text{fit\_quantizer}(\vec z, \beta_2)$
        \State $\vec s_q := \text{quantize}(\vec{s}, \vec s_s, \vec z_s)$
        \State $\vec z_q := \text{quantize}(\vec{z}, \vec s_z, \vec z_z)$
        \State $\mathcal{S} := \mathcal{S} \cup \{s_q, s_s, s_z, z_q, s_z, z_z\}$
        \State $\hat s := \text{dequantize}(s_q, s_s, s_z)$
        \State $\hat z := \text{dequantize}(z_q, z_s, z_z)$
      \State\Return $\hat s, \hat z, \mathcal{S}$
    \end{algorithmic}
\end{minipage}

\end{algorithm}

\textbf{Implementation details.} Our algorithm also contains several optimizations. 
As we are using small group sizes, it is often the case that a group contains all positive (or all negative) values. 
Standard quantizers~\cite{frantar2022obc, frantar2022gptq} require the maximum value to be positive and the minimum value to be negative. 
For small group sizes, removing this requirement results in slightly better quality. 
As a by-product of quantizing the quantization statistics, our algorithm allows non-integer zero points. 
We ablate these and other SpQR components in Section~\ref{sect:experiments}.

\subsection{Implementing and Leveraging the Sparse Quantized Representation}\label{sect:method_representation_and_inference}

\begin{figure}[t]
    \vspace{-0.5em}
    \centering
    \hspace{-10px}\includegraphics [width=400px] {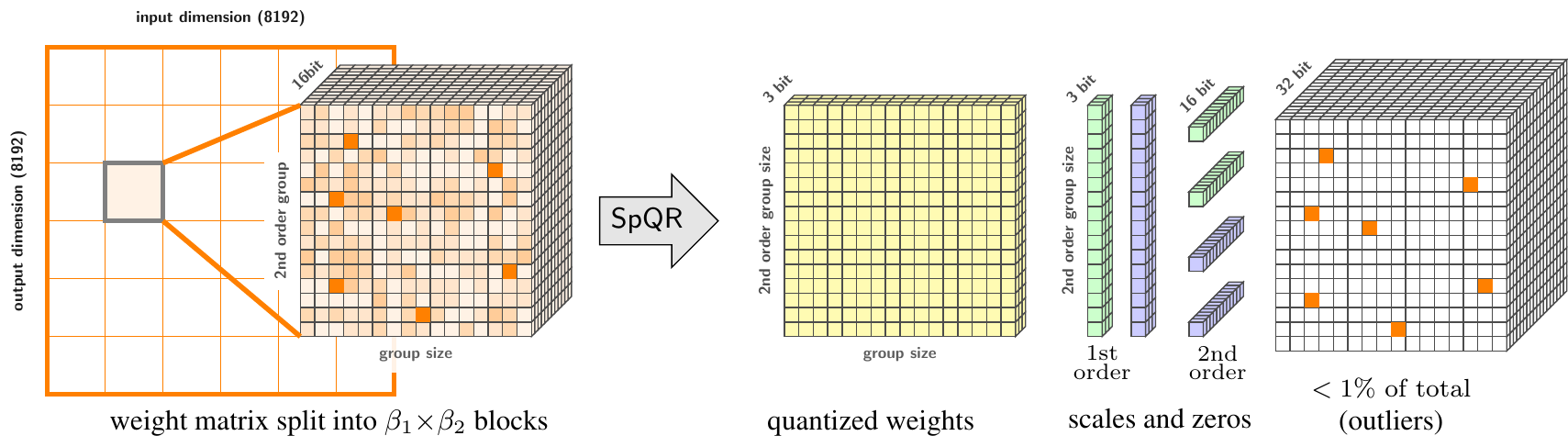} 
    \vspace{-8px}
    \caption{A high-level overview of the SpQR representation for a single weight tensor. The right side of the image depicts all stored data types and their dimensions.}
    \label{fig:spqr}
\end{figure}

Our algorithm converts homogeneous weights into several data structures of various sizes and precisions. Overall, the representation consists of (1) quantized weights, (2) first level quantized quantization statistics, second level quantization statistics, and (3) the CSR outlier indices and values. We summarize the overall structure of SpQR in Figure~\ref{fig:spqr} and describe each component below.

\textbf{Storing quantized groups.} All non-outlier weights are encoded as a structure that contains: 
\begin{itemize}
    \item a $b_w$-bit individual weight;
    \item a $b_q$-bit scale and zero point for each group of size $B$; 
    \item $16$-bit statistics for quantizing groups of $B_q$ quantization scales and zero-points. 
\end{itemize}

As a particular example for a SpQR representation, consider $b_w {=} b_q {=} 3$ and $B_w=B_q=16$. The weight matrix is split into groups of $B_q \times B_w = 256$ weights. A group contains 256 individual $b_w=3$-bit codes. Every 16 weights use a separate 3-bit scale and zero-point. Finally, there are four 16-bit scalars for the entire group used for second level quantization.
To simplify GPU memory access, we keep the quantized values for outlier weights in place and adjust the 16-bit versions to compensate for that. We also store both quantized weights and quantized quantization statistics in a contiguous memory region for each group. When running on a different hardware (e.g. mobile CPUs), it is possible to further reduce the memory footprint by removing the quantized version of outliers. We leave this direction for future work.

\textbf{Storing outliers.} 
% n our analysis, we found multiple groups of highly sensitive weights with different structure, including full-column groups, horizontal groups of uneven width and individual outliers. To accommodate all possible structures, we encode outliers individually, in a row-wise arrangement similar to CSR~\cite{todo}.
Recall that our outliers are unstructured; for storage, we sort them by their row first and column second, so that outliers in the same row are contiguous in memory. 
For each outlier, we store two scalars: the 16-bit weight value and the 16-bit column index. For each row, we also store a single 32-bit number---the total number of outliers in the rows up to the current one for efficient inference.
This results in an average storage cost of 32.03 to 32.1 bits per sensitive weight. 
This could be reduced significantly by grouping outliers, which we leave as future work. 
% In theory, it is possible to reduce this overhead by grouping outliers

% it is possible to slightly reduce storage overhead by storing some outliers together, as groups. It is also possible to quantize outlier values in less than 16 bit. In this work, we do not explore this direction in order to keep the overall representation simple.

\textbf{Inference with SpQR.} 
To illustrate the practicality of our approach, we design an efficient GPU-based decoding implementation for the SpQR format, focused on the popular token-by-token LLM generation as a use-case.  

We leverage the fact that autoregressive inference on GPUs is memory-bound, so high compression rates can hide decoding overheads, to a significant extent. 
At a high level, our algorithm loads group statistics and the quantized weights into shared memory (SRAM), dequantizes to 16-bits, and then performs matrix multiplication with 16-bit inputs.
For handling outliers, we design a sparse matrix algorithm that takes advantage of outliers that occur in rows. Roughly, the algorithm works as follows

First, (1) we divide the matrix into equally sized blocks.
Then, each GPU core (thread block) (2) loads a large slice of outliers into shared memory (SRAM), and each GPU core (3) determines if outliers are part of the segment or not. The corresponding weights are (4) loaded from main memory; finally, the matrix multiplication is performed.

This algorithm essentially performs load balancing through steps (1-3), while step (4) tends to have contiguous memory access due to the row-like patterns for the outliers. We will show in Section~\ref{sect:experiments} that this custom approach is faster than the sparse matrix algorithms in PyTorch.

\section{Experimental Validation}\label{sect:experiments}
% \vspace{-0.5em}
% In line 165 We promised additional visualizations for quantization error patterns.
% \subsection{Weight sensitivity to quantization [analysis]}\label{sect:experiments_sensitivity}
% need to decide on terms: quantization error /  outlier / weight sensitivity 
% When applying quantization to LLMs’ weights we observed that quantization errors (defined in~\ref{sect:method_analysis})%as difference between original and quantized weights, normalized by inverse Cholesky hessian
% are distributed unevenly and form patterns in weight matrix visualizations. Here, we explore these patterns in more detail.

\paragraph{Experimental setup.}
We focus on three main settings: 1) evaluating what is the most compact representation with which SpQR can replicate the performance of a 16-bit model within 1\% perplexity, 2) controlling for the average number of bits per parameter across methods and assess the performance of SpQR compared to round-to-nearest and GPTQ baselines, 3) what is the best trade-off in terms of model size and performance. For these settings, we evaluate the full SpQR algorithm on publicly-available LLMs. We focus on the LLaMA $\{7,13,30,65\}$B model family~\cite{touvron2023llama} and Falcon$\{7,40\}$B model family~\cite{falcon2023}. We quantize LLaMa models using the RedPajama dataset and Falcon models on RefinedWeb dataset~\cite{refinedweb2023}, publicly-available replicas of the LLaMA and Falcon training data, respectively. In addition, we provide perplexity results for OPT models in Appendix~\ref{app:lossless_compression}.

We compare SpQR against two other post-training quantization schemes: GPTQ~\cite{frantar2022gptq} and simple rounding-to-nearest (RTN) quantization, which is used by most other LLM compression methods~\cite{dettmers2022llm, yao2022zeroquant}. 
Both baselines use 4-bit quantization since it provides the best quality to size trade-off~\cite{dettmers2022case}. For SpQR, we consider both 3-bit and 4-bit base quantization, though the resulting model size can be slightly larger due to the presence of outliers.

We evaluate quantized model performance by two metrics. 
Firstly, we measure \emph{perplexity}, measured on the WikiText2~\cite{wikitext103}, Penn Treebank~\cite{PTB} and C4~\cite{C4} datasets. 
Secondly, we measure zero-shot accuracy on five tasks: WinoGrande~\cite{DBLP:journals/cacm/winogrande2021}, PiQA~\cite{tata2003piqa}, HellaSwag, ARC-easy and ARC-challenge~\cite{arc_allenai}. We use the LM Evaluation Harness~\cite{eval-harness} with recommended parameters. 
We provide full configurations in Appendix~\ref{app:configs}, as well as code which we plan to release publicly. 
Our implementation takes around 4.5 hours on the largest model size (65B) on an NVIDIA A100 and about 6 on an A6000. 

To control for model size, we evaluate RTN and GPTQ with 4-bit base quantization. 
For SpQR we use 3-bit base quantization, a group size of 8 with 3-bit for the first quantization, a group size of 64 for the second quantization, and as many outliers as possible to still reach less than 4-bits per parameter on average.
We aim to achieve \emph{near-lossless} compression, for which we adopt the definition of the MLCommons benchmark~\cite{reddi2020mlperf}: 
1\% error relative to the uncompressed baseline. In all SpQR evaluations, we choose $\tau$ such that the proportion of outliers is under $1\%$.

% While there is no universal criterion of what constitutes near-lossless compression, we follow the definition of popular benchmarks including MLCommons which recognize 1\% relative error to be insignificant for tasks need to reproduce target performance~\cite{dans_link_mlperf}. 

\paragraph{Main Results.}
Figure~\ref{fig:quantization_method_comparison} measures actual model size versus perplexity on LLaMa models on WikiText2, and accuracy on zero-shot tasks. 
We observe that SpQR outperforms GPTQ (and correspondingly RTN) at similar model size by a significant margin, especially on smaller models. 
This improvement comes from both SpQR achieving more compression, while also reducing loss degradation. 
In addition, if we measure the bits per parameter needed to come within 1\% of the 16-bit performance in terms of perplexity, Figure~\ref{fig:quantization_method_comparison} shows that SpQR with 4.6 to 4.71 bits per parameter approaches the non-quantized models with at most 1\% margin of error for all models (see Table~\ref{tab:quantization_method_comparison_LLaMA} and Table~\ref{tab:quantization_method_comparison_falcon} for exact values).

The second set of results, presented in Table~\ref{tab:quantization_method_comparison_LLaMA} for LLaMa and Table~\ref{tab:quantization_method_comparison_falcon} for Falcon family models, controls model size by comparing SpQR and baseline methods with 4 bits per parameter. 
These results show that SpQR improves over previous methods, with the gap between SpQR and the next best method GPTQ being as large as the improvement of GPTQ over naive RTN. For 4-bit, SpQR halves the error relative to the 16-bit baseline compared to GPTQ.
% Applying SpQR to the largest LLaMA model takes at most 6 hours on a single RTX A6000 GPU, with slight variation depending on the chosen group size and $\tau$.

\begin{figure}
\vspace{-0.5em}
\footnotesize
% \scriptsize
\centering
\setlength\tabcolsep{4pt}
\begin{floatrow}
\capbtabbox{%
\renewcommand{\arraystretch}{1.15}

\begin{tabular}{lccccc}
 % \multicolumn{4}{c}{LLAMA}
  \multicolumn{2}{l}{\bf{LLaMa}}  & \multicolumn{4}{c}{}\\

 \toprule
 \bf{Size} & \bf{Method} & \bf{Avg bits} & \bf{Wiki2} & \bf{C4} & \bf{PTB}\\
 \midrule
 % \midrule

 \multirow{5}{*}{7B} & -- & 16.00 & 5.68 & 7.08 & 8.80\\
  & SpQR & 4.63 & 5.73 & 7.13 & 8.88\\\cmidrule{2-6}
 & RTN & 4 & 6.43 & 7.93 & 10.30\\
 & GPTQ & 4 & 6.13 & 7.43 & 9.27\\
 & SpQR & 3.94 & 5.87 & 7.28 & 9.07\\
 \midrule
 \multirow{5}{*}{13B} & -- & 16.00 & 5.09 & 6.61 & 8.07\\
  & SpQR & 4.63 & 5.13 & 6.64 & 8.13\\\cmidrule{2-6}
 & RTN & 4 & 5.55 & 6.98 & 8.65\\
 & GPTQ & 4 & 5.40 & 6.84 & 8.44\\
 & SpQR & 3.96 & 5.22 & 6.72 & 8.22\\
 \bottomrule
 \end{tabular}
 \hfill
 \begin{tabular}{lccccc}
\multicolumn{6}{c}{}\\
 \toprule
  \bf{Size} & \bf{Method} & \bf{Avg bits} & \bf{Wiki2} & \bf{C4} & \bf{PTB}\\
 \midrule
 \multirow{5}{*}{30B} & -- & 16.00 & 4.10 & 5.98 & 7.30 \\
  & SpQR & 4.69 & 4.14 & 6.01 & 7.33\\\cmidrule{2-6}
 & RTN & 4 & 4.57 & 6.34 & 7.75\\
 & GPTQ & 4 & 4.48 & 6.20 & 7.54\\
 & SpQR & 3.89 & 4.25 & 6.08 & 7.38\\
 \midrule
 \multirow{5}{*}{65B} & -- & 16.00 & 3.53 & 5.62 & 6.91\\
  & SpQR & 4.71 & 3.57 & 5.64 & 6.93 \\\cmidrule{2-6}
 & RTN & 4 & 3.87 & 5.85 & 7.17\\
 & GPTQ & 4 & 3.83 & 5.80 & 7.07\\
 & SpQR & 3.90 & 3.68 & 5.70 & 6.99\\
 \bottomrule
\end{tabular}
}
{
  \caption{Perplexity on WikiText2~\cite{wikitext103}, C4~\cite{C4} and Penn Treebank~\cite{PTB} for SpQR and round-to-nearest (RTN) and GPTQ baselines with LLaMa. We can see that SpQR reaches performances within 1\% of the perplexity with less than 4.71 bits per parameter. We also see that for 4-bits per parameter SpQR significantly improves on GPTQ with an improvement as large as the improvement from RTN to GPTQ. }%
  \label{tab:quantization_method_comparison_LLaMA}
}  
\end{floatrow}
\end{figure}

\begin{figure}
\vspace{-0.5em}
\footnotesize
% \scriptsize
\centering
\setlength\tabcolsep{4pt}
\begin{floatrow}
\capbtabbox{%
\renewcommand{\arraystretch}{1.15}

\begin{tabular}{lccccc}
 % \multicolumn{4}{c}{LLAMA}
  \multicolumn{2}{l}{\bf{Falcon}}  & \multicolumn{4}{c}{}\\

 \toprule
 \bf{Size} & \bf{Method} & \bf{Avg bits} & \bf{Wiki2} & \bf{C4} & \bf{PTB}\\
 \midrule
 % \midrule

 \multirow{5}{*}{7B} & -- & 16.00 & 6.59 & 9.50 & 9.90\\
  & SpQR & 4.44 & 6.64 & 9.58 & 9.97\\\cmidrule{2-6}
 & RTN & 4 & 8.73 & 12.56 & 13.76\\
 & GPTQ & 4 & 6.91 & 9.93 & 10.33\\
 & SpQR & 3.92 & 6.74 & 9.70 & 19.114\\
 \bottomrule
 \end{tabular}
 \hfill
 \begin{tabular}{lccccc}
\multicolumn{6}{c}{}\\
 \toprule
  \bf{Size} & \bf{Method} & \bf{Avg bits} & \bf{Wiki2} & \bf{C4} & \bf{PTB}\\
 \midrule
 \multirow{5}{*}{40B} & -- & 16.00 & 5.23 & 7.76 & 7.83 \\
  & SpQR & 4.46 & 5.26 & 7.79 & 7.86\\\cmidrule{2-6}
 & RTN & 4 & 6.52 & 9.76 & 10.63\\
 & GPTQ & 4 & 5.36 & 7.95 & 8.01\\
 & SpQR & 3.90 & 5.29 & 7.85 & 7.91\\
 \bottomrule
\end{tabular}
}
{
  \caption{Perplexity on WikiText2~\cite{wikitext103}, C4~\cite{C4} and Penn Treebank~\cite{PTB} for SpQR and round-to-nearest (RTN) and GPTQ baselines on Falcon model. We can see that SpQR reaches performances within 1\% of the perplexity with less than 4.5 bits per parameter. We also see that for 4-bits per parameter SpQR significantly improves on GPTQ with an improvement as large as the improvement from RTN to GPTQ.}%
  \label{tab:quantization_method_comparison_falcon}
}  
\end{floatrow}
\end{figure}

\begin{figure}
\footnotesize
\begin{floatrow}
\capbtabbox{%
\renewcommand{\arraystretch}{1.15}
% \floatsetup[table]{capposition=top}
{\caption{Perplexity for LLaMA-65B model.}%
\label{Tab:ablation}}
\begin{tabular}{lcccc}
 \toprule
 \bf{Name} & \bf{Wiki2} & \bf{C4} & \bf{PTB} & \bf{Avg bits} \\
 \midrule 
 Uncompressed & 3.53 & 5.62 & 6.91 & 16\\
 GPTQ (4 bit) & 3.83 & 5.80 & 7.07 & 4 \\
 % SpQR (3 bit)   & 3.71 & 5.731 & 7.013 & 3.628\\
 \midrule
 3-bit statistics & 3.74 & 5.73 & 7.02 & 3.63\\
 16-bit statistics & 3.84 & 5.83 & 7.12 & 3.67\\
 \midrule
 Round zero & 3.75 &  5.76 & 7.01 & 3.63\\
 w/o act order & 3.74 & 5.76 & 7.05
 & 3.63\\
 % w/o perchannel & inf & inf & inf & 3.625\\
 % True Sequential & 3.716 & 5.739
 % & 7.029 & 3.625\\

 \bottomrule
\end{tabular}
}

\ffigbox{%
  \includegraphics[width=0.9\linewidth]{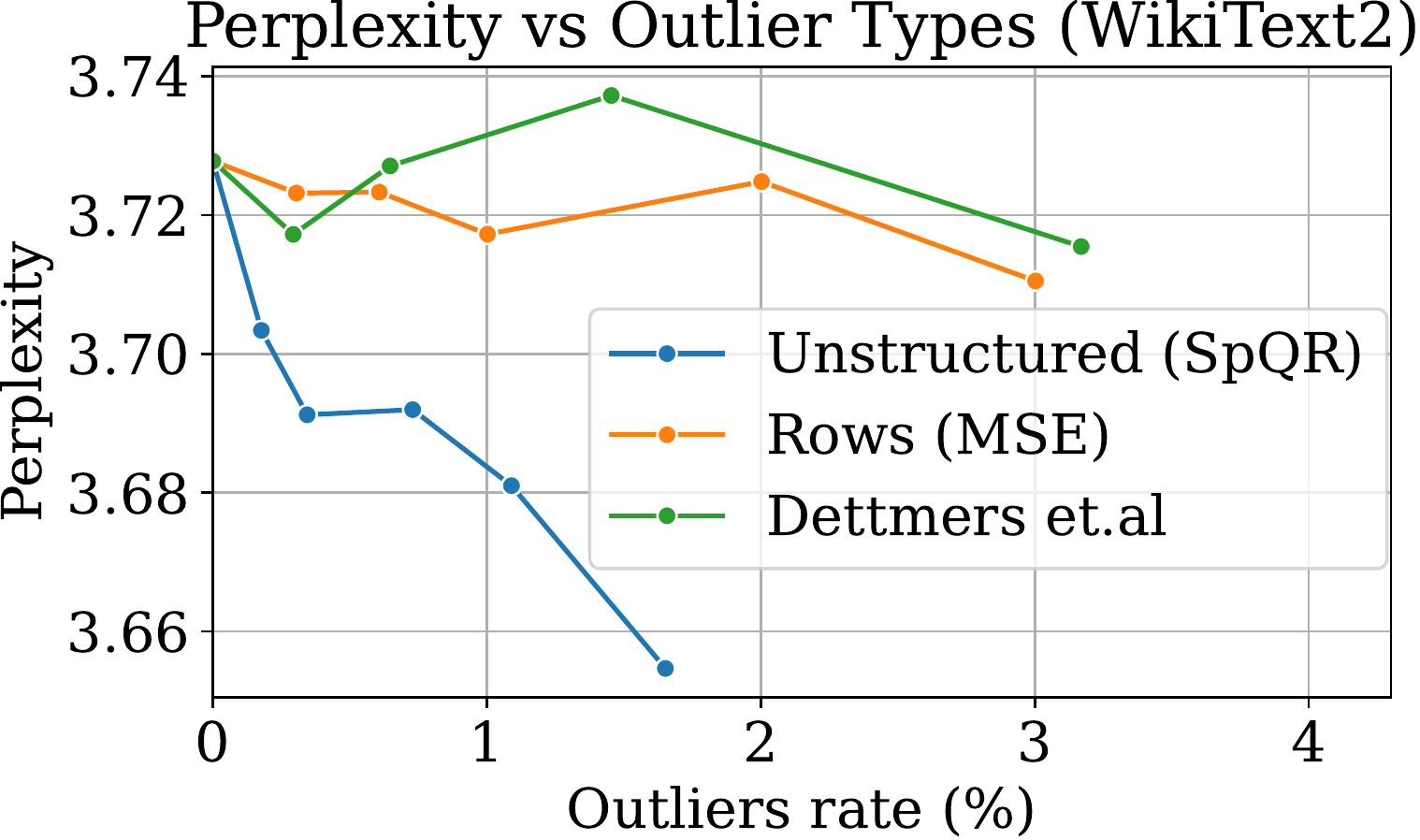}
}{%
  \caption{Different outlier types, LLaMA-65B. }%
  \label{fig:outliers_fig}
}
\end{floatrow}
\end{figure}

\paragraph{Ablations.}
The SpQR representation differs from standard quantization methods in two main ways: bilevel quantization with small quantization group size and unstructured outliers. To understand the effect of small group sizes, we compare 3-bit SpQR with group size 16, compressed using 3-bit bilevel quantization, versus a  setup with group size 48, keeping quantization statistics in 16-bit. Both configurations result in approximately 3.6 average bits per parameter. For simplicity, neither uses outliers.
We report both in Table~\ref{Tab:ablation}, the ``3-bit statistics`` entry corresponds to group size 16 with 3-bit statistics and ``16-bit statistics`` stands for group size 16 with 16-bit statistics. Given the same (slightly smaller) memory footprint, using quantized statistics significantly improves language modeling loss.

Next, we ask whether it is necessary to use unstructured outliers, considering two outlier types. First, we use the criterion of Dettmers et al.~\citep{dettmers2022case} to find column outliers and quantize them in higher precision. The alternative is to treat the entire rows (output units / hidden units / neurons) as outliers: we run SpQR without outliers, then select $k$ output units that have the highest quantization error (i.e. MSE between layer predictions) and treat the entire rows as 16-bit outliers. %We must also note that the unstructured outliers take up approximately twice as much space per outlier as row or column outliers.
We compare the three outlier types on top of 3-bit SpQR and report the results in Figure~\ref{fig:outliers_fig}. Overall, unstructured outliers reduce perplexity significantly faster than their row counterpart and the criterion of~\cite{dettmers2022case}, even after accounting for the different memory footprint.

Finally, we analyze the impact of the minor hyperparameter changes that we introduced at the end of Section~\ref{sect:method_compress}. In Table~\ref{Tab:ablation} (bottom), we evaluate quantization errors without these changes. The ``Round zero'' entry corresponds to a version of SpQR where the zero-point is a 3-bit integer. This reduces the memory footprint of SpQR, but results in a moderate increase in perplexity. Similarly, we evaluate SpQR without the ``act order'' flag. This option re-orders the input dimensions by the diagonal of the inverse hessian, which was introduced as a part of the GPTQ algorithm. Using this heuristic slightly improves loss, though not as much as from quantized groups.

To summarize, both small quantized groups and unstructured outliers independently improve perplexity and perform better than alternative strategies. SpQR also benefits from using the GPTQ activation order heuristic, though the gain is smaller than from outliers or small groups. Still, we opt to use the same activation order heuristic in the GPTQ baselines to ensure a fair comparison.
To further explore the design space of SpQR, we provide an additional hyperparameter study in Appendix~\ref{app:hyperparameter}.

% As we can see from Table \ref{Tab:ablation}, each of individual change gave positive change with small or none bits increase.
% As for Figure \ref{fig:outliers_fig} we can determine 

\paragraph{Inference Time.}\label{sect:experiments_speed}
Finally, we evaluate the inference speed of SpQR for autoregressive inference with a focus on measuring the token generation latency with batch size 1 on a single A100 GPU. We measure inference speed in two setups: i) generating 100 tokens from scratch and ii) adding 100 tokens on top of a 1024-token prefix (prompt). We compare our specialized sparse matrix multiplication algorithm with the algorithm implemented in PyTorch\nocite{pytorch} (cuSPARSE). We also compare against a 16-bit baseline. We measure the end-to-end latency as inference steps per second for the full SpQR algorithm, that is for both the dense and sparse multiplication part together.

\begin{table}[h]
    \vspace{-0.5em} % note to Dan: if my figure vspace  tweaks get in the way, tell me, ill stop
    \scriptsize
    \centering
    \setlength\tabcolsep{2pt}
    \begin{tabular}{c|cccc|cccc|cccc}
         Method        & \multicolumn{4}{c|}{fp16 (baseline)} & \multicolumn{4}{c|}{SpQR (PyTorch)} & \multicolumn{4}{c}{SpQR (optimized)} \\
         \midrule
         LLaMA         & 7B     & 13B     & 30B     & 65B     & 7B     & 13B     & 30B     & 65B      & 7B     & 13B     & 30B     &     65B \\
         scratch  & $47\pm2.3$ & $37\pm0.8$ & $19\pm1.1$ & OOM & $30\pm2.2$ & $24\pm1.2$ & $8.8\pm0.4$ & OOM & $\textbf{57}\pm2.4$ & $\textbf{44}\pm0.5$ & $\textbf{22}\pm0.9$ & $\textbf{12}\pm0.6$ \\
         prefix 1024   & $46\pm2.4$ & $31\pm0.9$ & $17\pm0.8$ & OOM & $27\pm1.6$ & $21\pm1.1$ & $6.5\pm0.7$ & OOM & $\textbf{55}\pm2.1$ & $\textbf{37}\pm0.8$ & $\textbf{22}\pm1.3$ & $\textbf{11}\pm0.6$ \\\bottomrule
    \end{tabular}
    \caption{Inference speed comparison (tokens/s), OOM means the model did not fit in an A100 GPU. We see that our optimized SpQR algorithm is faster than the 16-bit baseline and almost 2.0x faster than quantized matrix multiplication + standard PyTorch sparse matrix multiplication baseline.}
    \label{tab:inference}
\end{table}

Results are shown in Table~\ref{tab:inference}. We can see that while standard sparse matrix multiplication in PyTorch is not faster than 16-bit inference, our specialized sparse matrix multiplication algorithm yields speedups of about 20-30\%.
% This makes SpQR highly a highly compact representation that yields faster inference than regular 16-bit models.

\vspace{-0.7em}
\section{Discussion \& Limitations}
\vspace{-0.5em}

We have presented SpQR, an quantization approach which quantizes sensitive outliers in higher precision, to achieve near-lossless 16-bit accuracy with less than 4.75 bits per parameter on average. We achieve even better quality-size-tradeoff when compressing to as little as 3.36 bits which makes SpQR an ideal method for compressing models for memory-limited devices.
Despite our promising results, there are several limitations. The main limitation is that we do not evaluate the generative quality of quantized LLMs, but only the predictive performance in terms of zero-shot accuracy and perplexity. While we believe that perplexity measurements and generation quality are strongly related, this is a hypothesis we aim to investigate in future work. 
While we devise a sparse matrix multiplication algorithm to accelerate the computation with outliers, another limitation is that we do not fuse sparse matrix multiplication with regular quantized matrix multiplication. Such an approach would yield even better inference time performance. However, such an approach is also very difficult to implement. We leave the implementation of such an algorithm to future work.

\section{Acknowledgements}
D.K. was supported by Russian Science Foundation, grant 21-11-00373. 
D.A. and E.F. gratefully acknowledge funding from the European Research Council (ERC) under the European Union’s Horizon 2020 research and innovation programme (grant agreement No 805223 ScaleML). Authors also thank Ivan Komarov for his help in profiling and understanding the performance bottlenecks of SpQR on GPU. 

% A further limitation is that we discover structured highly sensitive outliers in the model weights, but we offer not explanation as to what these outliers are and what they represent. While these outliers seem to be important for LLM predictive performance it is unclear if they could be removed while preserving performance to enhance quantization precision. We leave this research to future work.

% \section{Broader Impact}

% Our method enables the deployment high quality LLMs in the 7-13B parameters range to memory-limited devices such as laptops and phones. With our method it is possible to develop specialized 7B LLMs in hassle-free 16-bit and then enable the deployment of such LLMs unto phones by applying SpQR. Since SpQR is practically lossless, this ensures a reliable level performance of deployed LLMs which is important for consumer applications. Since mobile phones are ubiquitous and LLMs powerful general purpose tools, SpQR might have a wide-reaching effect on how LLMs are used by the general population to complete useful tasks.

% Since LLMs are an inherent dual-use technology that can be used for either good or bad, the total impact for users is undetermined. However, we believe that the impact will be largely positive since users usually can chose which third-party applications are installed on their phone and as such they can choose to install appliations that are beneficial to them.

%%%%%%%%%%%%%%%%%%%%%%%%%%%%%%%%%%%%%%%%%%%%%%%%%%%%%%%%%%%%

\bibliography{main}
\bibliographystyle{alpha}

\pagebreak
% - for each mention of appendix / supplementary in the paper, please create a section here
% --- note that not all of them have \appendix  tag
% - add a \label{app:something} with a *best effort* meaningful name
% - the title and/or comment should reflect what we promised to write there
% - the order of sections should be the same as in the paper, i.e. the first appendix metnion should refer to A, then B, etc.
\appendix
\renewcommand{\contentsname}{Table of contents}
{\small\setlength{\parskip}{0.05em}\tableofcontents}
\section{Additional weight sensitivity analysis}\label{app:extra_analysis}
% note: consider adding pagebreak; optional though; why: to avoid appendix A having the first few lines on a separate page

In this section, we provide additional visualizations of LLaMA weight sensitivities, as well as additional plots for different layer roles.
As we observed earlier in Section~\ref{sect:method_llama65}, the sensitivity matrices vary based on four main factors:
\begin{itemize}
    \item the quantization scheme (e.g. row- or group-wise);
    \item the layer depth, i.e. the index of the corresponding transformer block;
    \item the role of that weight, e.g. self-attn query / key or MLP up / down projection;
    \item the location within the chosen weight matrix;
\end{itemize}

Here, we report additional observations about these factors and elaborate on some of our claims from Section~\ref{sect:method_understand}.
We also report raw sensitivity matrices for various weight matrices at the end of the supplementary materials.

\paragraph{Relation between sensitivity and the chosen quantization scheme.} We compare two configurations of GPTQ 3-bit. The first configuration uses one quantization scale \& zero for each row. The second one uses blockwise quantization with one set of statistics for each block of 128 weights.

\begin{figure}[h]
    \centering
    \includegraphics[width=0.9\linewidth]{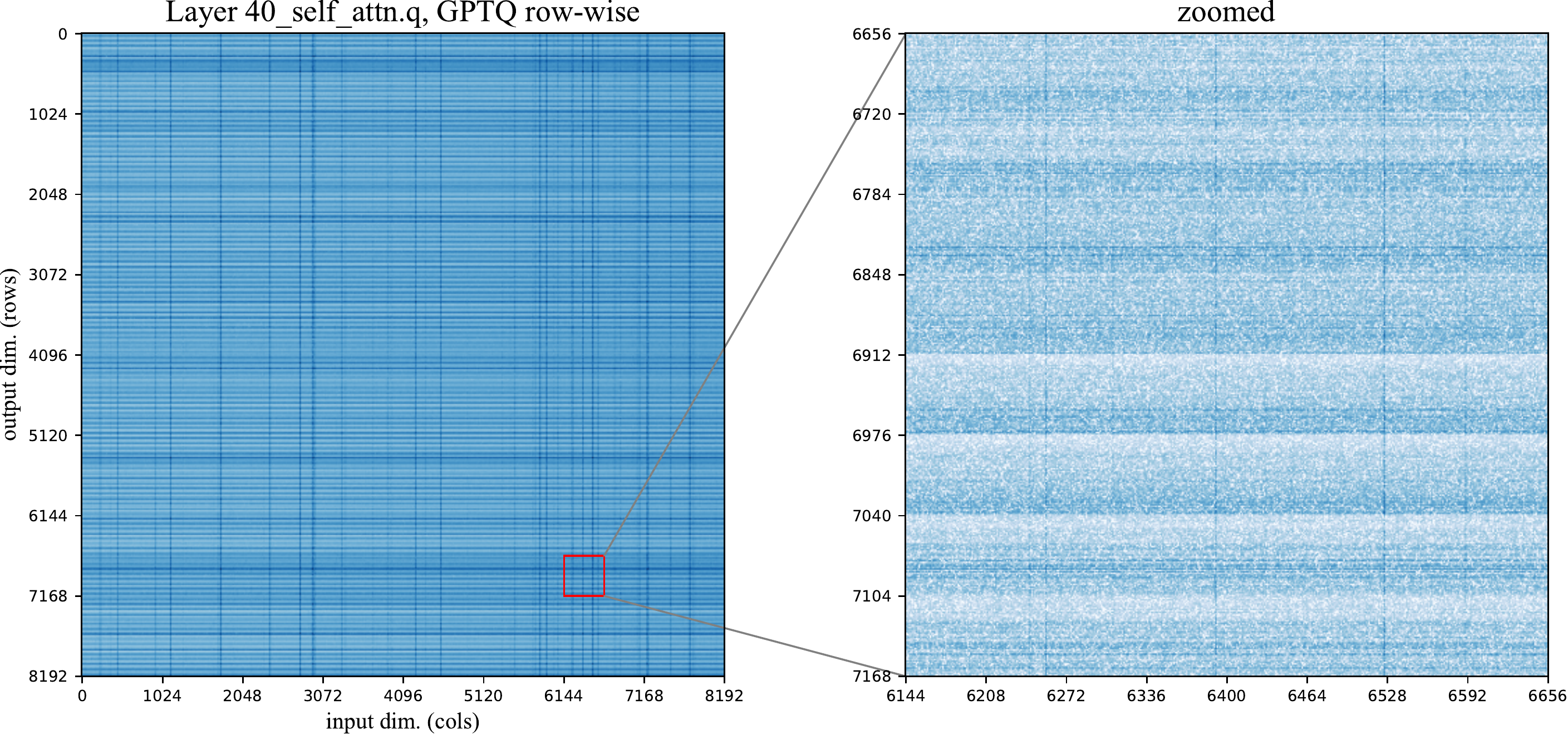}
    \includegraphics[width=0.9\linewidth]{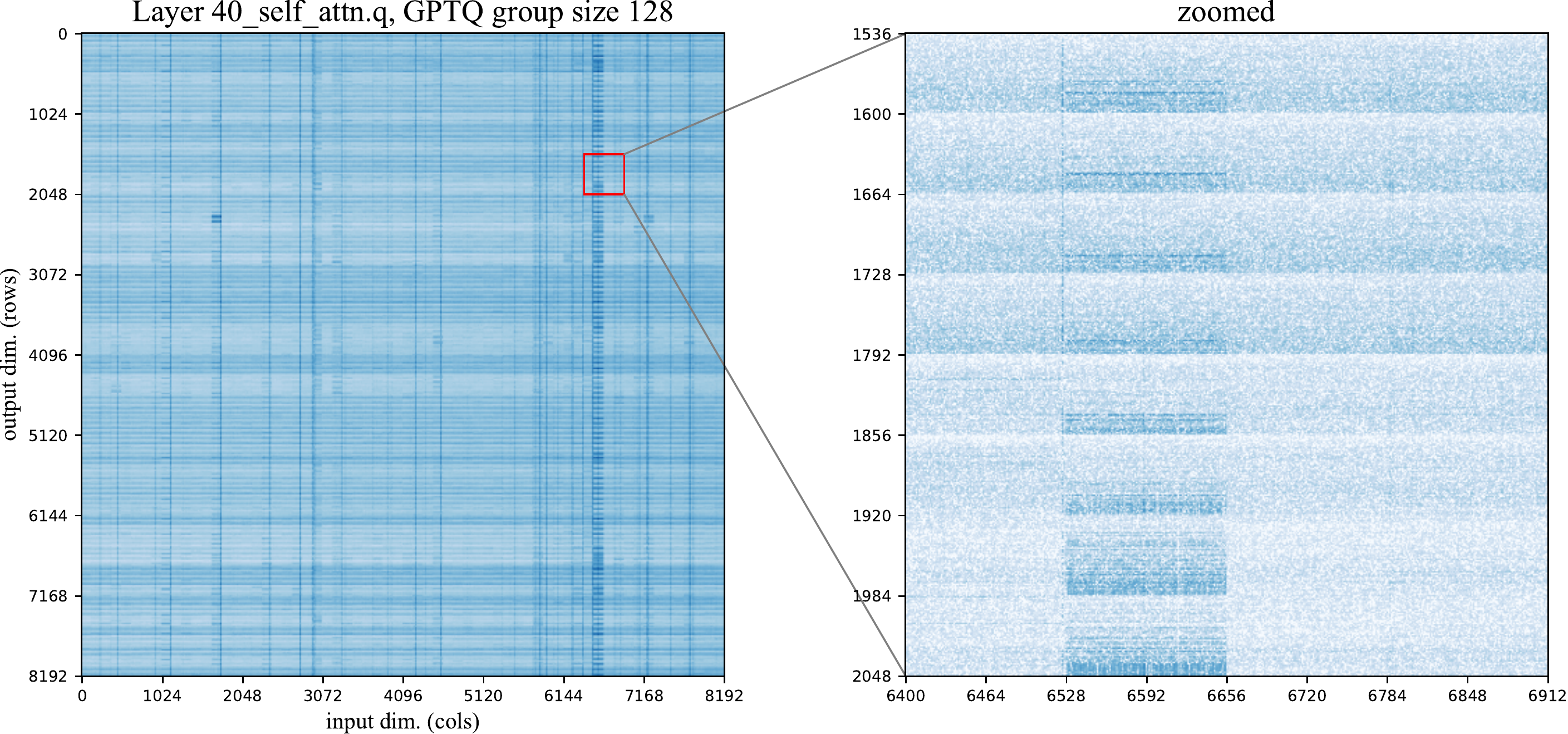}
    \caption{The weight sensitivities for LLaMA-65B 40th layer, attention query projection. 
    The color scale represents sensitivity on a logarithmic scale, with higher sensitivity being darker. \textbf{(top)} 3-bit GPTQ with per-row quantization scales, \textbf{(bottom)} 3-bit GPTQ with block size 128.}
    \label{fig:analysis_sensitivity_block128}
\end{figure}

Figure~\ref{fig:analysis_sensitivity_block128} demonstrates a typical example of how group size affects sensitivity. In the bottom-right plot, we observe that a subset of weights (width 128) has a significantly higher quantization error than the rest of the layer. Please note that the color scale represents sensitivity on a logarithmic scale, with higher sensitivity being darker.

On a more detailed examination, we found that this specific group contains a ``vertical'' outlier, i.e. the corresponding input feature has significantly higher variance, compared to other input dimensions.

In this example, the main effect of GPTQ block size 128 is that the problematic dimension leads to increased sensitivity in a group of $8192 \times 128$ weights. In turn, GPTQ with per-row statistics has high quantization error across the entire row.

\textbf{The effect of rotary embeddings.} Earlier in Figure~\ref{fig:patterns} we note that attention query and key have a regular pattern of sensitivity that repeats every 64 rows. We attribute this to the fact that LLaMA uses rotary position embeddings. More specifically, this pattern is likely a side-effect of how rotary embeddings are implemented for this model.

To recall, rotary position embeddings are a technique that rotates attention head dimensions by an angle that depends on how many tokens are between key and query~\cite{su2021roformer}. Furthermore, dimensions within each head are rotated with a different frequency. To implement this rotation, LLaMA multiplies each head by a precomputed tensor of sine and cosine functions with a different period. The first half (64 units) of the matrix is multiplied by cosines and the other half (64 units) is multiplied by sines.

To recall, sine and cosine components are equivalent up to a phase shift and show similar behavior in our analysis. In general, we observe that weights that correspond to low-frequency heads (bottom of each semi-head) typically have higher sensitivity. One possible explanation is that high-frequency heads can be more dependent on position-specific information, such as attending to the previous token --- and less dependent on the weights that represent content information. However, this phenomenon merits further investigation and our current understanding should be treated as an educated guess.

\textbf{GPTQ and the effect of quantization order.} As we observe earlier in Section~\ref{sect:method_llama65}, the rightmost weights in each visualization tend to have higher quantization errors. This is likely a side-effect of the GPTQ algorithm, which compresses weights one input feature at a time, i.e. column by column in a left-to-right direction. Once a column is quantized, the algorithm uses the remaining unquantized weights to compensate for the error. Thus, the rightmost batch of weights accumulates the most error from preceding columns and has the least space to compensate it's ``own'' quantization error.

\begin{figure}[h]
    \centering
    \includegraphics[width=0.7\linewidth]{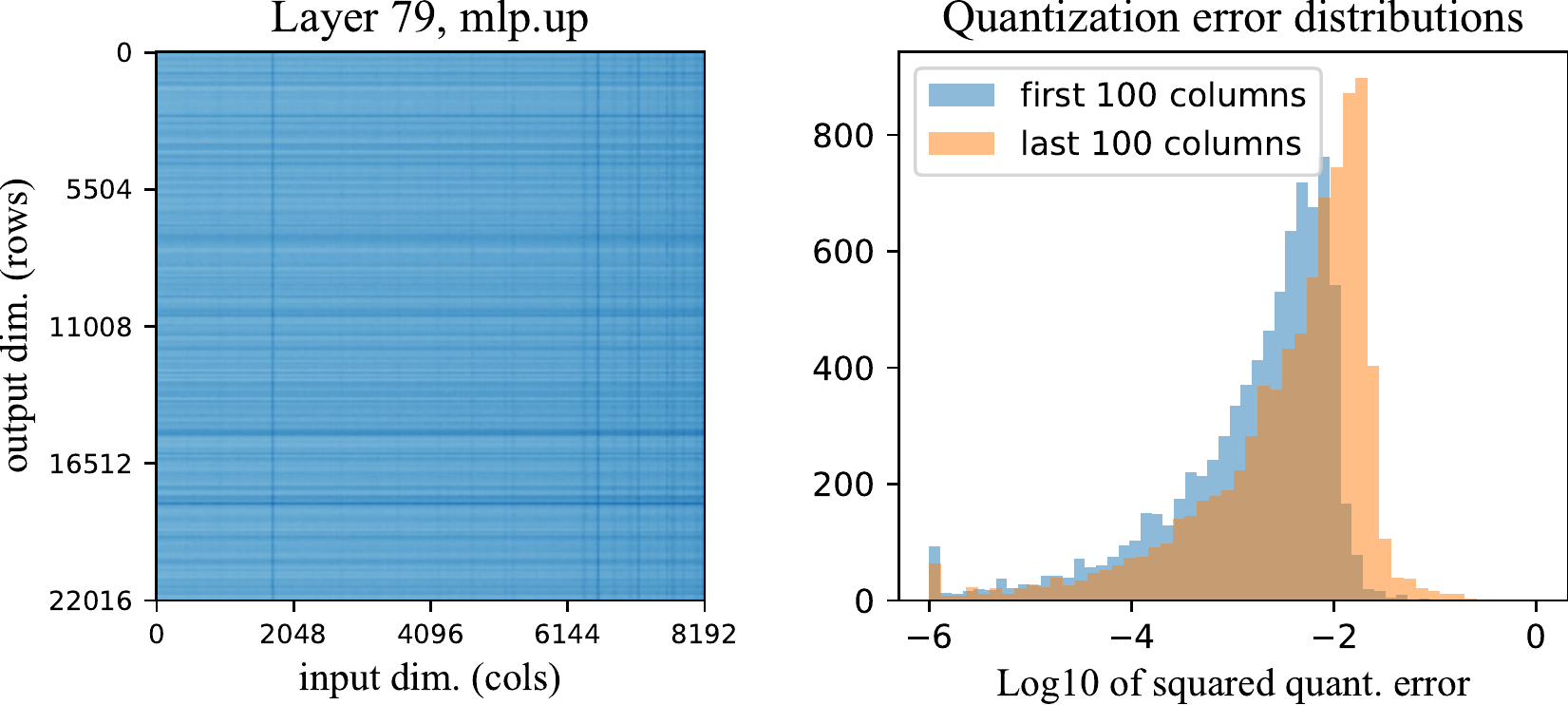}
    \caption{The weight log-sensitivities for a deeper upward projection layer (in particular, this is layer \#79). The heatmap on the left represents the sensitivities of each weight, with darker being more sensitive; the histogram on the right captures the sensitivities in the first 100 and last 100 columns (sorted across input dimensions). The latter figure clearly shows that later columns are more sensitive on average.}
    \label{fig:sensitivity_left2right}
\end{figure}

This difference is most pronounced in the earlier layers, where the quantization error is smaller overall (see Figure~\ref{fig:sensitivity_left2right}). To further verify this observation, we observe that this effect disappears if we shuffle the weight quantization order in the GPTQ algorithm.

\begin{figure}[h]
    \centering
    % \vspace{-15px}
    \includegraphics
        [width=\linewidth]
        {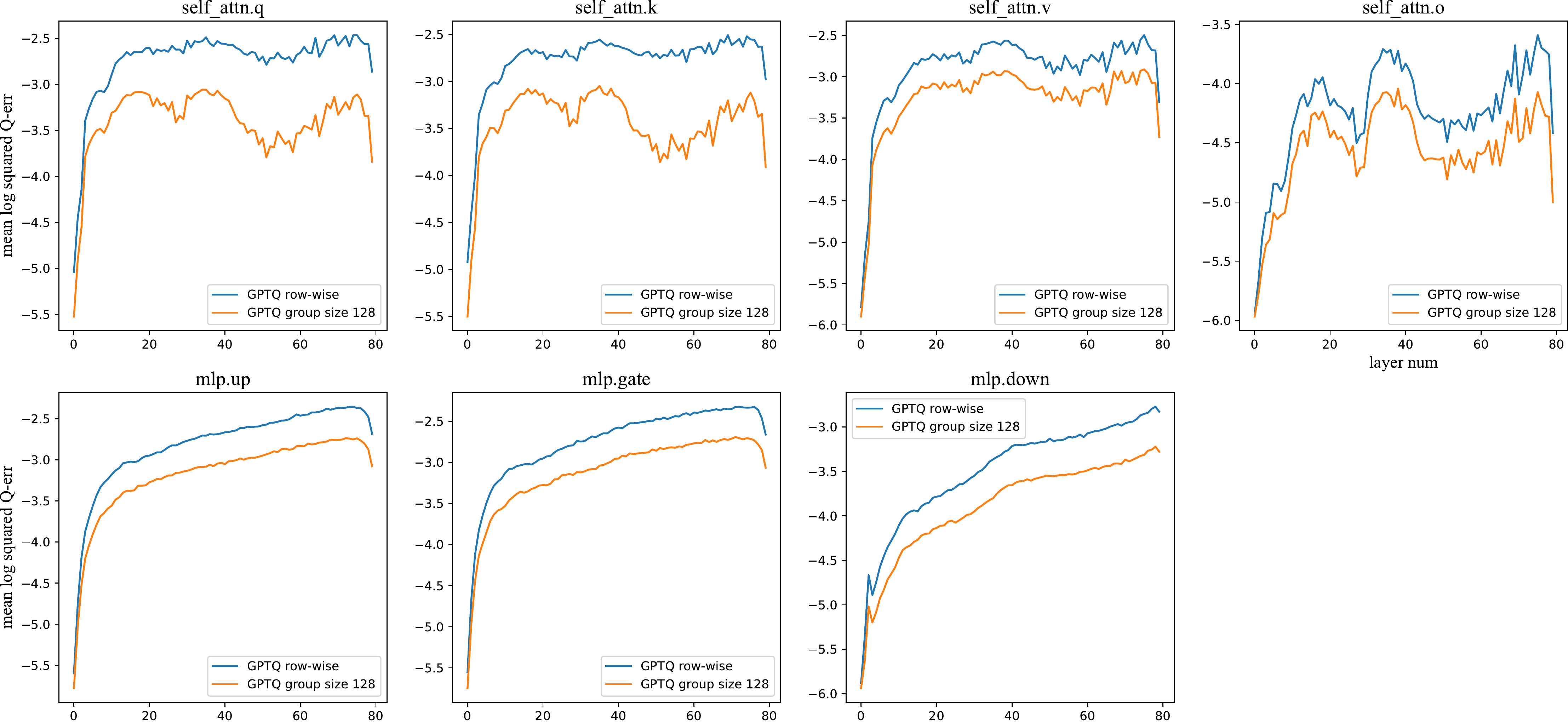}
    \caption{Figure: mean quantization error (vertical axis) as a function of layer depth (horizontal axis). Each plot corresponds to a different layer role.}
    \label{fig:error_by_depth}
\end{figure}

\textbf{Relation between weight sensitivity and layer depth.} 
In terms of mean squared error, we observe that the first layers of LLaMA tend to have generally lower OBC error (defined as L2 distance between original and quantized layer predictions).
% However, the difference in OBC error does not guarantee that these layers are easier to quantize or less ``important''.
%Instead, early layers seem to have generally lower output variance. 
To illustrate this, we report the average quantization error of GPTQ-3bit in Figure~\ref{fig:error_by_depth}.

\begin{figure}[h!]
    \centering
    \vspace{-20px}
    \includegraphics
        [height=21.5cm]
        % [width=\linewidth * 0.5]
        {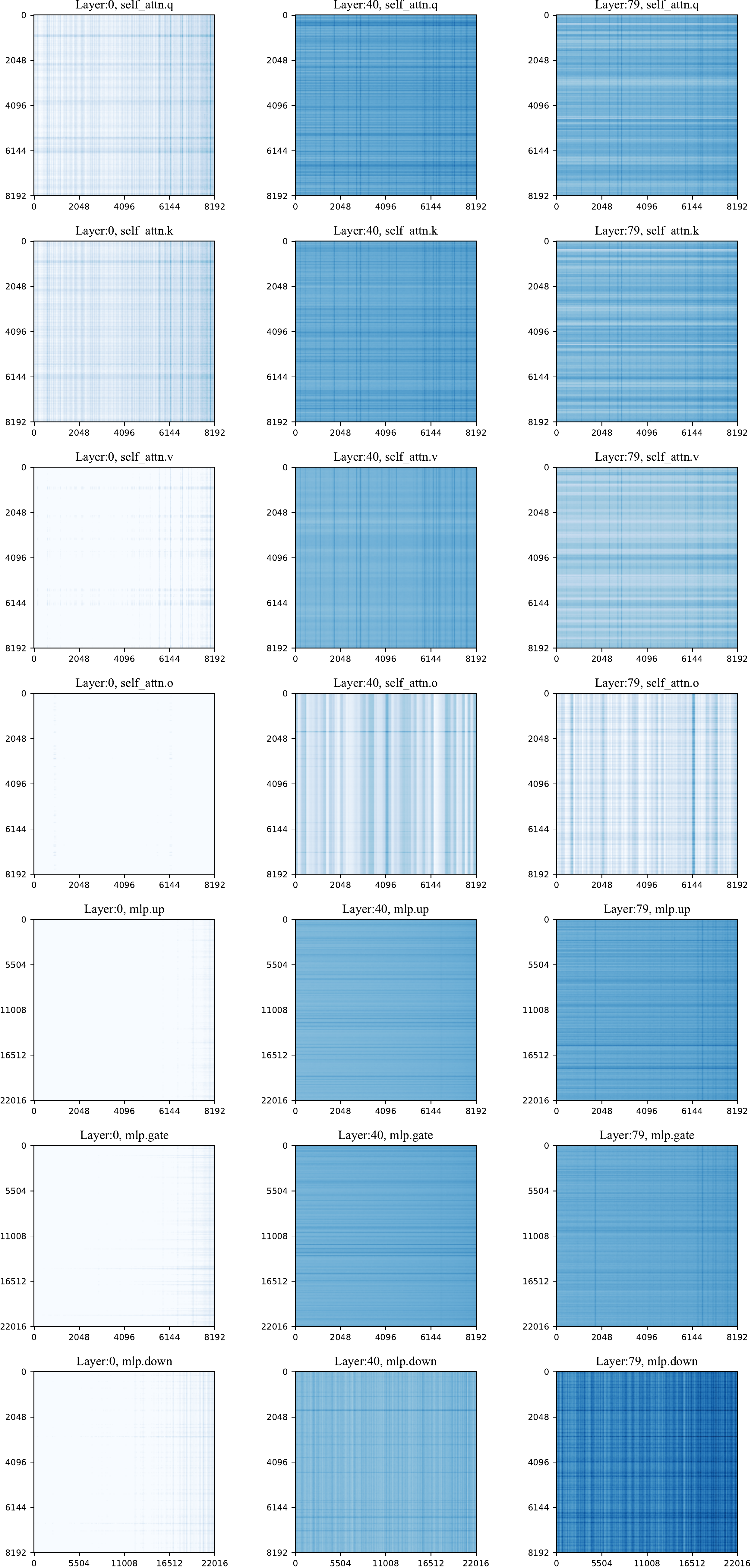}
    \vspace{-5px}
    \caption{A grid of weight log-sensitivities for LLaMA-65B for 3-bit GPTQ compression with per-row quantization statistics. Each row corresponds to a specific layer type (e.g. attention query, mlp gate), and the columns represent layer depth.}
    \label{fig:sensitivities_3x7_noblock}
\end{figure}

\begin{figure}[h!]
    \centering
    \vspace{-20px}
    \includegraphics
        % [width=\linewidth]
        [height=21.5cm]
        {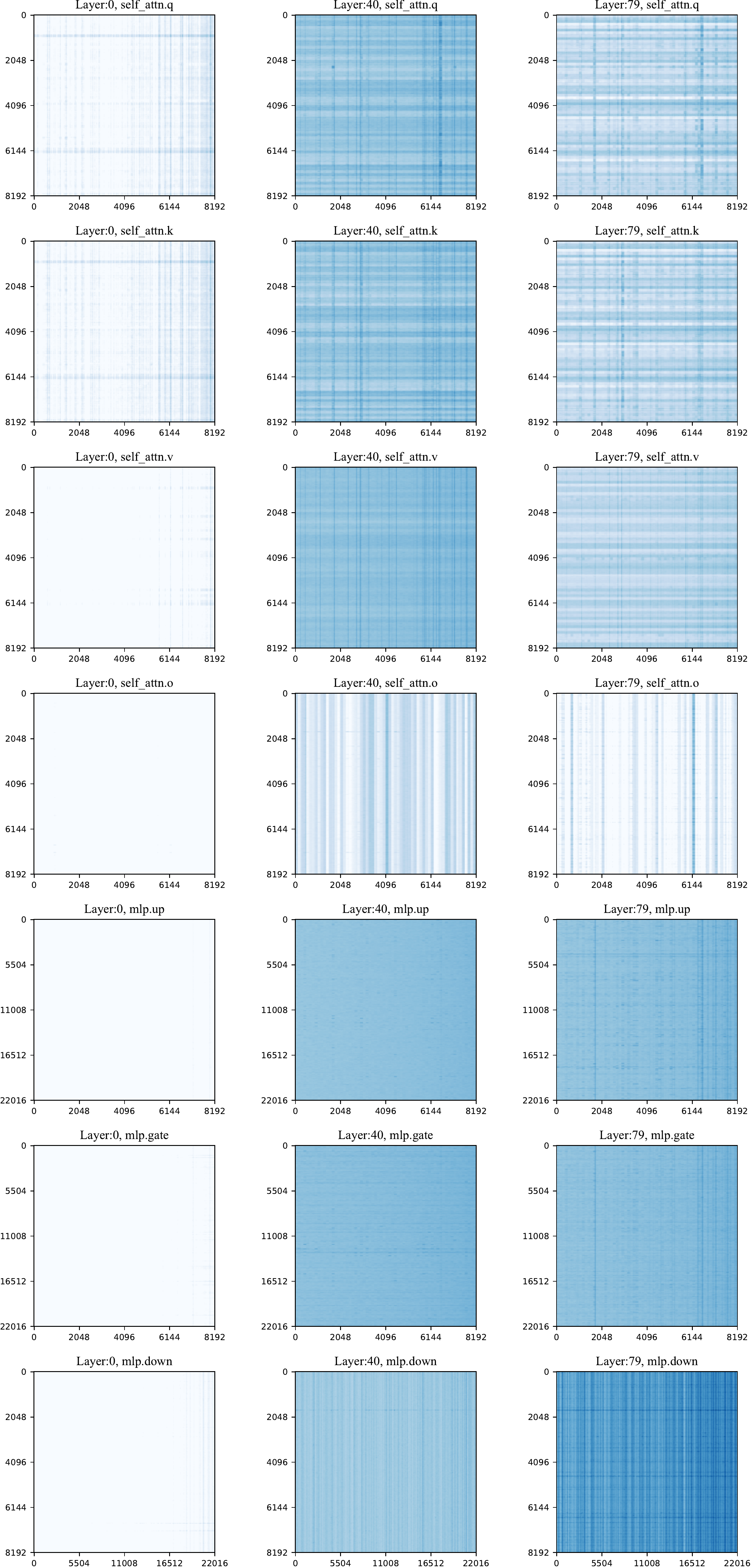}
    \vspace{-5px}
    \caption{A grid of weight log-sensitivities for LLaMA-65B for 3-bit GPTQ compression with group-wise quantization of block size 128. Each row corresponds to a specific layer type (e.g. attention query, mlp gate), and the columns represent layer depth.}
        \label{fig:sensitivities_3x7_block128}

\end{figure}

The absolute quantization error means little by itself since each quantized layer has a different input/output variance. However, we also observe that the first and last few layers have qualitative differences in behavior. Figures~\ref{fig:sensitivities_3x7_noblock} and~\ref{fig:sensitivities_3x7_block128} report weight sensitivities for the first, middle (40th), and last (79th) layer of LLaMA model separately to better illustrate this difference.

\section{Experimental Configurations}\label{app:configs}

The SpQR representations proposed in this work have several adjustable hyperparameters that allow for great flexibility in targeting a desired size of the model. 
We introduce the notation and list the method hyperparameters below:

\begin{itemize}
    \item $b_{w}$ - number of bits per weight
    \item $b_{s}$ - number of bits per scale
    \item $b_{z}$ - number of bits per zero
    % \item $B$ - size of block for quantization
    % \item $B_q$ - size of the block for quantization of quantized statistics (scales and zeros)
    \item $r_o$ - outlier rate (fraction of weights that are not quantized) 
    \item $\beta_1$ - block size for weight quantization
    \item  $\beta_2$ - block size for statistic quantization;
    \item $\tau$ - outlier threshold
\end{itemize}

The actual number of outliers depends not only on $\tau$, but on all other hyperparameters as well. However, for any specific configuration, increasing $\tau$  leads to reduced number of outliers. To achieve the desired number of outliers, we tune $\tau$ in $[0.1, 1.0]$ range by binary search with minumum step size $0.05$. The vast majority of our configurations are between $\tau=0.1$ and $\tau=0.45]$.

The full configuration we use to compress LLaMA-30B model near-losslessly in Table~\ref{tab:quantization_method_comparison_LLaMA} has the following hyperparameters: $b_w=4, b_s=b_z=3, \beta_1=\beta_2=16, \tau=0.1$ This translates to the following command line arguments in our supplementary code:

\begin{verbatim}
python main.py $MODEL custom --custom_data_path=$DATA  \
    --wbits 4 --groupsize 16 --perchannel --qq_scale_bits 3  \
    --qq_zero_bits 3 --qq_groupsize 16 --outlier_threshold 0.1 \
    --fit_quantizer_without_outliers --permutation_order act_order  
\end{verbatim}

\section{Hyperparameter sensitivity}\label{app:hyperparameter}

%In 342 line We promised to provide additional hyper parameter study.

In this section, we analyze how SpQR performance depends on the choice of quantization group sizes.
Please recall that the SpQR algorithm uses two types of groups, indexed by parameters $\beta_1$ and $\beta_2$.
The first group dimension $\beta_1$ covers multiple weights for the same input unit, similar to standard blockwise quantization. In turn, the other dimension $\beta_2$ covers multiple output units, and is used when quantizing quantization scales. In our visualizations, $\beta_1$ blocks are always horizontal, while $\beta_2$ are vertical. 

In Table~\ref{tab:groupsize_analys}, we evaluate SpQR with varying parameters $\beta_1$ and $\beta_2$. We quantize LLaMA-65B with 3-bit SpQR for weights and statistics and report perplexity on WikiText2, Penn Treebank, and C4 datasets. The upper-left section of the table contains the effective number of bits for each group configuration, and the remaining sections correspond to perplexities on different datasets.
\begin{table}[h!]
    \vspace{-0.5em} 
    \normalsize
    \centering
    \setlength\tabcolsep{4pt}
    \begin{tabular}{c|cccccc|cccccc|}
                 & \multicolumn{6}{c|}{\textbf{Average bits}} & \multicolumn{6}{c|}{\textbf{Wikitext2 Perplexity (3.53)}}  \\
         \midrule
         \backslashbox{$\beta_1$}{$\beta_2$}         & 4     & 8     & 16     & 32     & 64     & 128     & 4     & 8      & 16     & 32     & 64     &     128 \\

          \midrule
          4 & 8.5	& 6.5 & 5.5 & 5 & 4.75 & 4.625 & 3.581 & 3.628 & 3.715 & 3.822 & 4.003 & 4.23\\
          8 & 5.75 & 4.75 & 4.25 & 4 & 3.875 & 3.813 & 3.625 & 3.64 & 3.649 & 3.666 & 3.688 & 3.713\\
          16 & 4.375 & 3.875 & 3.625 & 3.5 & 3.438 & 3.406 & 3.701 & 3.71 & 3.728 & 3.726 & 3.739 & 3.741\\
          32 & 3.688 & 3.438 & 3.313 & 3.25 & 3.219 & 3.203 & 3.803 & 3.797 & 3.812 & 3.812 & 3.815 & 3.85\\
          64 & 3.344 & 3.219 & 3.156 & 3.125 & 3.109 & 3.102 & 3.884 & 3.901 & 3.907 & 3.899 & 3.928 & 3.926\\
          \vspace{1 em}
          128 & 3.172 & 3.109 & 3.078 & 3.063 & 3.055 & 3.051 & 3.982 & 3.994 & 4.005 & 3.992 & 4.017 & 4.013\\

        % \midrule
         & \multicolumn{6}{c|}{\textbf{C4 Perplexity (5.62)}} & \multicolumn{6}{c|}{\textbf{PTB Perplexity (6.91)}}  \\
         \midrule
          \backslashbox{$\beta_1$}{$\beta_2$}     & 4     & 8     & 16     & 32     & 64     & 128     & 4     & 8      & 16     & 32     & 64     &     128 \\
          \midrule
          4 & 5.652 & 5.674 & 5.718 & 5.796 & 5.919 & 6.119 & 6.934 & 6.965 & 7.001 & 7.054 & 7.194 & 7.395\\
          8 & 5.683 & 5.688 & 5.696 & 5.703 & 5.709 & 5.718 & 6.962 & 6.98 & 6.991 & 6.99 & 6.979 & 7.029\\
          16 & 5.735 & 5.735 & 5.735 & 5.738 & 5.741 & 5.749 & 7.018 & 7.013 & 7.015 & 7.016 & 7.012 & 7.03\\
          32 & 5.793 & 5.789 & 5.792 & 5.796 & 5.794 & 5.802 & 7.042 & 7.053 & 7.083 & 7.043 & 7.069 & 7.083\\
          64 & 5.857 & 5.859 & 5.858 & 5.866 & 5.863 & 5.866 & 7.084 & 7.129 & 7.137 & 7.118 & 7.137 & 7.12\\
          128 & 5.932 & 5.931 & 5.935 & 5.939 & 5.944 & 5.936 & 7.185 & 7.197 & 7.232 & 7.234 & 7.217 & 7.199\\
    \end{tabular}
    \caption{Weight block size $\beta_1$ and statistic block size $\beta_2$  performance on WikiText2, C4, and Penn Treebank (PTB). The uncompressed baseline value is provided in the corresponding heading.}
    \label{tab:groupsize_analys}
\end{table}

\section{Estimating model size}\label{app:modelsize}

% We compare the performance of 
% different quantization methods and choices of hyperparameters in 
In this section, we provide a quick way to estimate the compressed model size before running the quantization. We express this estimate in
terms of \emph{average bits per parameter} defined as:
\begin{equation}
\overline{b} = 
\frac{\mathrm{model\ size\ in\ bits}}{\mathrm{number\ of\ parameters}}
\end{equation}

Where $\mathrm{model\ size\ in\ bits}$ denotes the total amount of memory - the 
quantized weights, 1st-order and 2nd-order quantization statistics, outliers and the outlier index - required
for the storage of the model. According to Section \ref{sect:method_representation_and_inference}, 
each outlier requires memory storage of $\sim 32$ bits. 

% Naturally, outliers also affect the resulting number of bits. Using the storage method described in Section~\ref{sect:method_representation_and_inference}, each outlier takes up slightly less than 32.1 bits of storage. As a result, a model with 0.3\% outliers will have $\approx 0.16$ extra bits per parameter.

%%% Explain the quantization scheme. 

The storage and computational cost in transformer models are dominated by the linear projections
in the attention and feedforward blocks. Consider quantization of a weight matrix (any of these) $\mathbb{R}^{d_{\mathrm{out}} \times d_{\mathrm{in}}}$ with input dimension $d_{\mathrm{in}}$
and output dimension $d_{\mathrm{out}}$. Then the average number of bits for a given configuration is:

\begin{equation}
\overline{b} \simeq 
\frac{
b_{w} d_{\mathrm{out}} d_{\mathrm{in}} + 
(b_{s} + b_{z}) \frac{d_{\mathrm{out}} d_{\mathrm{in}}}{\beta_1} + 
2 (16 + 16) \frac{d_{\mathrm{out}} d_{\mathrm{in}}}{\beta_1 \beta_2}
}
{d_{\mathrm{out}} d_{\mathrm{in}} } + 32 r_o = 
b_{w} + \frac{b_{s} + b_{z}}{\beta_1} + \frac{64}{\beta_1 \beta_2} + 32 r_o
\end{equation}

Therefore, to increase (decrease) the size of the model one 
should either increase (decrease) the precision of model weights and quantization statistics
or decrease (increase) the block size.

For example, for configuration with $b_w = 3, b_s=3, b_z=3, \beta_1=16, \beta_2=32$ and $0.4\%$ of outliers, the average number of bits is:

$$
3 + \frac{3 + 3}{16} + \frac{64}{16 \cdot 32} + 0.004 \cdot 32 \simeq 3.63
$$

\section{Choice of optimal configuration for fixed average number of bits}

\begin{figure}[h]
    \centering
    \includegraphics[width=0.8\linewidth]{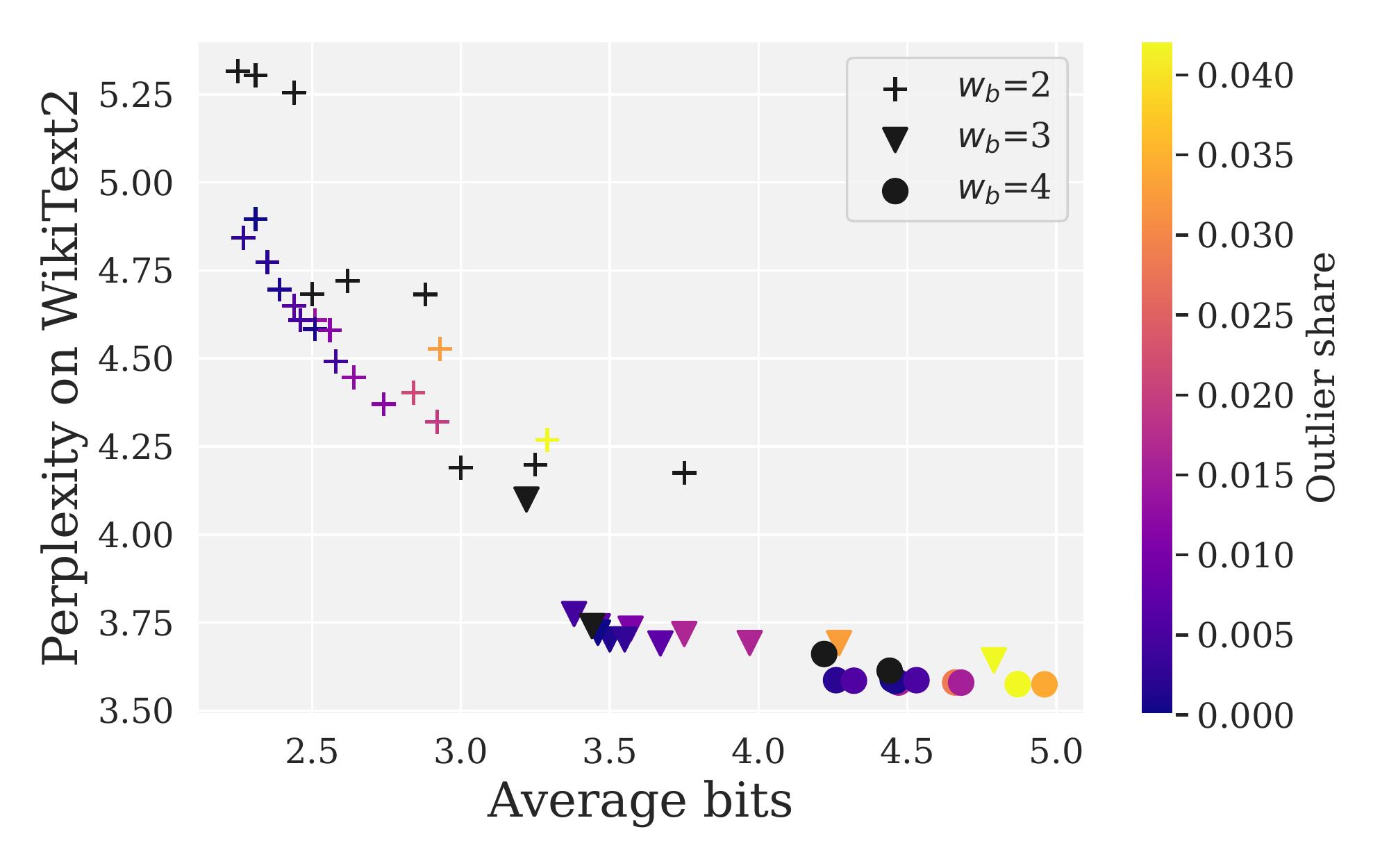}
    \caption{Perplexity of WikiText2 vs average number of bits. Different markers denote different $b_{w}$. Black colors correspond to quantization configurations without outliers and the brightness of the color is proportional to the outlier rate. }
    \label{fig:wikitext2_vs_wbits_avg_vs_outlier_shape}
\end{figure}

As discussed above our method has multiple options for improvement of model performance at the cost of the 
increase of the model size: number of bits per weight $w_b$, groupsizes $b_1$ and $b_2$ for 1st and 2nd order quantization and the
outlier rate. We evaluated several configurations with various options for the aforementioned parameters on perplexity benchmarks. 
Results are presented on Figure \ref{fig:wikitext2_vs_wbits_avg_vs_outlier_shape}. 
One can observe that small groups and small fraction of outliers allows to considerably improve model performance, 
but the gain is diminishing with the number of bits added (when the additional budget from small group is of order 0.1-0.5 of bits per parameter).
It is better to store weights in higher precision instead of keeping them in lower precision but with very small groups or 
keeping large fraction of outliers. In our experiments optimal fraction of outliers is 0.2-0.5\% depending on the model and groupsize.

\section{Additional results for near-lossless compression}\label{app:lossless_compression}

In this section we report the list of quantization configurations for OPT in Table \ref{tab:quantization_method_comparison_OPT}  on WikiText2, Penn Treebank, and C4 datasets. 

In addition we report results for LM eval harness for LLaMa Table \ref{tab:quantization_method_comparison_LM_eval_LLaMA}.
and recently released \href{https://falconllm.tii.ae}{Falcon} models - Falcon-7B and Falcon-40B Table \ref{tab:quantization_method_comparison_LM_eval_Falcon}.

\begin{figure}
\vspace{-0.5em}
\footnotesize
% \scriptsize
\centering
\setlength\tabcolsep{4pt}
\begin{floatrow}
\capbtabbox{%
\renewcommand{\arraystretch}{1.15}

\begin{tabular}{lccccc}
 % \multicolumn{4}{c}{LLAMA}
  \multicolumn{2}{l}{\bf{OPT}}  & \multicolumn{4}{c}{}\\

 \toprule
 \bf{Size} & \bf{Method} & \bf{Avg bits} & \bf{Wiki2} & \bf{C4} & \bf{PTB}\\
 \midrule
 % \midrule

 \multirow{5}{*}{6.7B} & -- & 16.00 & 10.86 & 11.74 & 13.09\\
  & SpQR & 4.27 & 10.81 & 11.88 & 13.17\\\cmidrule{2-6}
 & RTN & 4 & 12.10 & 13.38 & 16.09\\
 & GPTQ & 4 & 11.39 & 12.15 & 13.80\\
 & SpQR & 3.94 & 11.04 & 11.98 & 13.33\\
 \midrule
 \multirow{5}{*}{13B} & -- & 16.00 & 10.12 & 11.20 & 12.34\\
  & SpQR & 4.27 & 10.22 & 11.27 & 12.41\\\cmidrule{2-6}
 & RTN & 4 & 11.32 & 12.35 & 15.4\\
 & GPTQ & 4 & 10.31 & 11.36 & 12.58\\
 & SpQR & 3.93 & 10.28 & 11.34 & 12.52\\
 \bottomrule
 \end{tabular}
 \hfill
 \begin{tabular}{lccccc}
\multicolumn{6}{c}{}\\
 \toprule
  \bf{Size} & \bf{Method} & \bf{Avg bits} & \bf{Wiki2} & \bf{C4} & \bf{PTB}\\
 \midrule
 \multirow{5}{*}{30B} & -- & 16.00 & 9.56 & 10.69 & 11.84\\
  & SpQR & 4.26 & 9.50 & 10.73 & 11.88\\\cmidrule{2-6}
 & RTN & 4 & 10.97 & 11.90 & 14.17\\
 & GPTQ & 4 & 9.63 & 10.80 & 11.98\\
 & SpQR & 3.94 & 9.54 & 10.78 & 11.93\\
 \midrule
 \multirow{5}{*}{66B} & -- & 16.00 & 9.33 & 10.28 & 11.36\\
  & SpQR & 4.23 & 9.37 & 10.32 & 11.40\\\cmidrule{2-6}
 & RTN & 4 & 110 & 249 & 274\\
 & GPTQ & 4 & 9.55 & 10.50 & 11.58\\
 & SpQR & 3.91 & 9.32 & 10.35 & 11.43\\
 \bottomrule
\end{tabular}
}
{
  \caption{Perplexity on WikiText2~\cite{wikitext103}, C4~\cite{C4} and Penn Treebank~\cite{PTB} for SpQR and round-to-nearest (RTN) and GPTQ baselines with OPT. We can see that SpQR reaches performances within 1\% of the perplexity with less than 4.3 bits per parameter. We also see that for 4-bits per parameter SpQR significantly improves on GPTQ with an improvement as large as the improvement from RTN to GPTQ.}%
  \label{tab:quantization_method_comparison_OPT}
}  
\end{floatrow}
\end{figure}

\begin{figure}
\vspace{-0.5em}
\footnotesize
\centering
\setlength\tabcolsep{4pt}
\begin{floatrow}
\capbtabbox{%
\renewcommand{\arraystretch}{1.15}
\begin{tabular}{lcccccccc}
  \multicolumn{2}{l}{\bf{LLaMA}}  & \multicolumn{7}{c}{}\\
 \toprule
 \bf{Size} & \bf{Method} & \bf{Avg bits} & \bf{Winogrande} & \bf{Piqa} & \bf{Hellaswag} & \bf{Arc easy} & \bf{Arc challenge} & \bf{Avg score}\\
 \midrule
															
 \multirow{5}{*}{7B} & -- & 16.00 & 67.09 & 78.32 & 56.41 & 67.38 & 38.23 & 61.492\\
  & SpQR & 4.63 & 67.48 & 78.45 & 56.01 & 67.13 & 38.23 & 61.460\\\cmidrule{2-9}
 & RTN & 4 & 64.72 & 76.44 & 53.49 & 63.51 & 36.60 & 58.952\\
 & GPTQ & 4 & 65.35 & 77.58 & 54.99 & 63.55 & 36.35 & 59.564	\\
 & SpQR & 3.45 & 67.48 & 78.13 & 55.27 & 65.87 & 38.05 & 60.960\\
 \midrule
 \multirow{5}{*}{13B} & -- & 16.00 & 70.09 & 78.89 & 59.11 & 74.54 & 43.94 & 65.314\\
  & SpQR & 4.63 & 69.77 & 78.94 & 59.02 & 74.37 & 43.17 & 65.054\\\cmidrule{2-9}
 & RTN & 4 & 69.61 & 78.24 & 57.34 & 72.56 & 42.58 & 64.066\\
 & GPTQ & 4 & 69.06 & 78.40 & 58.04 & 73.23 & 43.26 & 64.398\\
 & SpQR & 3.45 & 68.90 & 78.73 & 58.22 & 73.27 & 42.75 & 64.374\\
  \midrule
 \multirow{5}{*}{30B} & -- & 16.00 & 72.93 & 80.96 & 62.66 & 75.34 & 46.76 & 67.730\\
  & SpQR & 4.69 & 72.93 & 81.01 & 62.50 & 76.05 & 47.18 & 67.934\\\cmidrule{2-9}
 & RTN & 4 & 72.06 & 79.05 & 60.61 & 70.66 & 42.24 & 64.924\\
 & GPTQ & 4 & 72.61 & 79.92 & 61.07 & 71.8 & 44.28 & 65.936\\
 & SpQR & 3.49 & 73.32 & 80.47 & 61.96 & 74.75 & 46.93 & 67.486\\
 \midrule
 \multirow{5}{*}{65B} & -- & 16.00 &77.43 & 81.50 & 63.95 & 75.17 & 47.10 & 69.030\\
  & SpQR & 4.71 & 76.95 & 81.56 & 63.76 & 75.25 & 46.93 & 68.890 \\\cmidrule{2-9}
 & RTN & 4 & 75.14 & 81.45 & 62.79 & 72.64 & 44.97 & 67.398\\
 & GPTQ & 4 & 75.85 & 80.79 & 62.91 & 74.20 & 46.59 & 68.068\\
 & SpQR & 3.52 & 76.09 & 81.18 & 63.54 & 74.37 & 45.05 & 68.046\\
 \bottomrule
 \end{tabular}

}
{
  \caption{LM eval harness results on LLaMA models.}%
  \label{tab:quantization_method_comparison_LM_eval_LLaMA}
}  
\end{floatrow}
\end{figure}

\begin{figure}
\vspace{-0.5em}
\footnotesize
\centering
\setlength\tabcolsep{4pt}
\begin{floatrow}
\capbtabbox{%
\renewcommand{\arraystretch}{1.15}
\begin{tabular}{lcccccccc}
  \multicolumn{2}{l}{\bf{Falcon}}  & \multicolumn{7}{c}{}\\
 \toprule
 \bf{Size} & \bf{Method} & \bf{Avg bits} & \bf{Winogrande} & \bf{Piqa} & \bf{Hellaswag} & \bf{Arc easy} & \bf{Arc challenge} & \bf{Avg score} \\
 \midrule								
 \multirow{5}{*}{7B} & -- & 16.00 & 67.32 & 79.49 & 57.77 & 74.71 & 40.1	0 & 63.878 \\
 & SpQR & 4.44 & 67.09 & 79.16 & 57.21 & 73.86 & 38.99 & 63.262 \\\cmidrule{2-9}
 & RTN & 4.00 & 65.51 & 77.37 & 51.86 & 68.69 & 33.7	& 59.426 \\
 & GPTQ & 4.00 & 66.38 & 79.11 & 56.68 & 73.15 & 38.48 & 62.760 \\
 & SpQR & 3.49 & 67.88 & 79.54 & 57.08 & 74.03 & 39.08 & 63.522 \\
 \midrule
 \multirow{5}{*}{40B} & -- & 16.00 & 76.62 & 82.32 & 64.06 & 82.03 & 50.26 & 71.058 \\
 & SpQR & 4.46 & 76.48 & 82.1 & 63.8 & 81.78	& 50.77 & 70.986 \\\cmidrule{2-9}
 & RTN & 4.00 & 75.69 & 80.30 & 60.52 & 79.92 & 49.83 & 69.252 \\
 & GPTQ & 4.00 & 75.93 & 81.23 & 63.05 & 80.85 & 50.00 & 70.212 \\
 & SpQR & 3.45 & 76.32 & 81.77 & 63.70 & 81.10 & 49.83 & 70.544 \\
 \bottomrule
 \end{tabular}
}
{
  \caption{LM eval harness results on Falcon models.}%
  \label{tab:quantization_method_comparison_LM_eval_Falcon}
}  
\end{floatrow}
\end{figure}

%      \\\\\         \\\\\
%     \\\\\\\__o    \\\\\\\__.
%   __\\\\\\\'/_____\\\\\\\'/___
% this area is infested with hedgehogs; proceed with caution

\section{Choice of optimal LLM configuration for specific hardware}\label{app:hardware_considerations}

In the preceding discussion, we were searching for optimal model configuration given some 
compression target without targeting any specific hardware or device. However, the 
question practitioner willing to deploy a model for a specific application would ask is:
What is the best model and compression setup for a given memory constraint? 

In this section, we provide a list of recommendations for the choice of the best LLaMA model and the corresponding 
compression level
that fits into the device memory (RAM or VRAM) without the need of offloading model parameters and activations.
We cover a range of available budgets from mobile devices to high-end workstation GPUs.
Recommendations are presented in Table~\ref{tab:hardware_considerations}.

\begin{table}[h!]
\begin{center}
\begin{tabular}{lccc}
\toprule
Device & Memory (GiB) & LLaMA & $\overline{b}$ \\
\midrule
iPhone13 & 4 & 7B & $\leq 3.5$ \\
\midrule
\multirow{2}{*}{iPhone14} & \multirow{2}{*}{6} & 7B & $\simeq 4.5$  \\
 & & 13B & $\leq3.5$ \\
\midrule
Consumer laptop & 8 & 13B & $\leq 4$ \\
\midrule
RTX4070 & 10-12 & 14B & $\simeq 4.5$ \\
\midrule
RTX4080 & 16 & 30B & $\leq 4$ \\
\midrule
RTX4090 & 24 & 30B & $\simeq 4.5$ \\
\midrule
V100 & 32 & 65B & $\leq 3.5$ \\
\midrule
A6000 & 48 & 65B & $\simeq 4.5$ \\
\bottomrule
\end{tabular}
\caption{Choice of the best LLaMA for a given memory constraint.}
\label{tab:hardware_considerations}
\end{center}
\end{table}

\section{Sensitivity to random seed}\label{app:randomseed}

The experiments we report throughout Section~\ref{sect:experiments} use one fixed random seed (the default value from the supplementary code). To verify that our results are robust to randomness, we run SpQR with 5 random seeds (0-5) and measure the adjusted standard deviation.

For this evaluation, we compress LLaMA-65B with SpQR using $b_w=b_z=b_s=3$ and $\beta_1=\beta_2=16$, which corresponds to 3.625 bits per parameter. The resulting perplexity scores are $3.75\pm0.003$ (WikiText2), $7.03\pm0.01$ (Penn Treebank) and $5.75\pm0.00086$ (C4). In addition to the chosen random seed, these standard deviations can be affected by the inherent nondeterminism of GPU computation. Overall, the standard deviations are at least one order of magnitude smaller than the difference between SpQR, GPTQ, and RTN.

\section{Generative examples}\label{app:examples}
Finally, we showcase several examples of how SpQR quantization affects the generated samples. For this evaluation, we take several prompts and use the compressed language model to continue generating text from these prompts. We compare the original LLaMA-65B and two quantized versions: SpQR and RTN-4bit. More specifically, we use the SpQR configuration that corresponds to near-lossless compression from Table~\ref{tab:quantization_method_comparison_LLaMA}. We use greedy autoregressive inference for all generated samples to ensure reproducibility.
The examples in Figure~\ref{fig:examples_txt_gen_4} show that all models produce a valid text, but SpQR matches the 16-bit model more frequently. The near-lossless algorithm also seems to produce more semantically similar texts. % However, it is unclear whether or not there is a statistically significant difference.

\begin{figure} { \tt \footnotesize \begin{tabularx}{\linewidth}{r X} \toprule Prompt: &  Every time data scientist Kim ran her new algorithm, it kept outputting the same unexpected result: a recipe for \\ 
 \midrule 16bit: & iced tea.  Kim was working on a project to predict the likelihood of a patient being readmitted to the hospital within 30 days of discharge. She was using a machine learning algorithm to analyze the data, and the algorithm  \\ 
 
 SpQR: &  iced tea.  Kim was working on a project to predict the likelihood of a patient being readmitted to the hospital within 30 days of discharge. She had been given a dataset of 100,000\\ 
 
 RTN 4-bit: & iced tea.  “I’m not sure what’s going on,” she said. “I’ve been running it for days, and it keeps giving me the same result.”  “I’m not sure what’s going on
\\  \end{tabularx} }  
{ \tt \footnotesize \begin{tabularx}{\linewidth}{r X} \toprule Prompt: &  Data scientist Grace was working late one night when she discovered a glitch in the matrix of her data. It spelled out \\ 
 \midrule 16bit: & 100\% proof that the world was going to end in 10 days.  She tried to tell her boss, but he didn’t believe her.  She tried to tell her friends, but they didn’t believe her  \\ 
 
 SpQR: &  100\% proof that the world was going to end in 10 days.  She tried to tell her boss, but he just laughed and said, “I’m not going to believe in some data glitch.”  She\\ 
 
 RTN 4-bit: &   100\% in binary code.  She was so excited she ran to tell her boss, but he was busy.  She ran to tell her colleague, but he was busy.  She ran to tell her friend, but she
\\ \end{tabularx} } \\ { \tt \footnotesize \begin{tabularx}{\linewidth}{r X} \toprule Prompt: &  Last Tuesday, just as the sun was setting, my neighbor's dog, Rover, started to speak to me in perfect French. He said  \\ 
 \midrule 16bit: & "Je suis un chien." I was shocked. I said, "Moi aussi."  I'm not sure what to make of this. I'm not sure what to make of anything.  I'm not sure what  \\ 
 
 SpQR: &  "Je suis un chien." I was shocked. I said, "Rover, you can speak French?" He said, "Oui, je parle francais." I said, "How did you learn to speak French?"\\ 
 
 RTN 4-bit: &   "Bonjour, Monsieur. Je m'appelle Rover. Je suis un chien. Je suis un chien français. Je suis un chien français qui parle français. Je suis un chien français qui parle français
\\  \end{tabularx} }  
 { \tt \footnotesize \begin{tabularx}{\linewidth}{r X} \toprule Prompt: &  In a parallel universe where cats are the rulers of Earth, our feline king was pondering his first decree \\ 
 \midrule 16bit: & " to make all dogs illegal.  "I"m not sure that"s a good idea," said his chief adviser, a cat named Whiskers.  "Why not?" asked the king.  "Because,  \\ 
 
 SpQR: &  " to make all dogs illegal.  "I"m not sure that"s a good idea," said his chief adviser, a cat named Whiskers.  "Why not?" asked the king.  "Because,\\ 
 
 RTN 4-bit: &  " to make the world a better place for cats.  He was about to sign the decree when he was interrupted by a knock on the door.  "Come in," he said.  The door opened and a cat entered.
\\ \bottomrule \end{tabularx} }  \caption{Texts generated by different quantized LLaMA-65B models with the same prompt.}  \label{fig:examples_txt_gen_4}  \end{figure}%%%----------

\section{Broader impact}\label{app:broader_impact}
Our method enables the deployment of high-quality LLMs in the 7-13B parameters range to memory-limited devices such as laptops and phones. With our method, it is possible to develop specialized 7B LLMs in hassle-free 16-bit and then enable the deployment of such LLMs to phones by applying SpQR. Since SpQR is practically lossless, this ensures a reliable performance level for deployed LLMs which is important for consumer applications. Since mobile phones are ubiquitous and LLMs powerful general-purpose tools, SpQR might have a wide-reaching effect on how LLMs are used by the general population to complete useful tasks.

%Since LLMs are an inherent dual-use technology that can be used for either good or bad, the total impact on users is undetermined. 
LLMs are inherently a dual-use technology that can bring both significant benefits and serious harm. The ethical and societal risks of LLMs range from deliberate malicious use (e.g. generating spam) and accidental misuse to adverse economic side-effects~\cite{weidinger2021ethical}.
% However, we believe that the impact will be largely positive since users usually can choose which third-party applications are installed on their phone and as such they can choose to install applications that are beneficial to them.
However, we believe that the marginal impact of SpQR will be positive or neutral since the LLMs we use are already openly available. Better quantization algorithms like SpQR let users with low-end devices run larger and generally more accurate language models. In other words, our algorithm does not create models with new capabilities (and risks): it only makes existing models more accessible.

\section{On the use of LLMs in this work}
Following the request in this year's call for papers, we describe the use of large language models in our paper. We used two different chat-based language models: ChatGPT and Claude+. We used these models to accelerate the process of writing LaTeX code in Alg.~\ref{alg:main} and Figure~\ref{fig:spqr} (via Tikz). We also used these LLMs to provide slight improvements to the table design throughout the paper.

In addition to this, we use ChatGPT to generate some prompts for Appendix~\ref{app:examples}. Finally, we used Claude+ to produce possible formulations for the outlier criterion in Alg.~\ref{alg:main}. In all these cases, we used LLMs through chat-based user interfaces, instructing them to generate code (LaTeX) or suggest improvements. If the suggested changes would not work as expected, we reported them to the model in natural language, using the same chat-based interface.

\end{document}